\documentclass[letterpaper]{article} 
\usepackage{aaai2026}  
\usepackage{times}  
\usepackage{helvet}  
\usepackage{courier}  
\usepackage[hyphens]{url}  
\usepackage{graphicx} 
\usepackage{natbib}  
\usepackage{caption} 
\usepackage{algorithm}
\usepackage{algorithmic}
\usepackage{amsfonts}
\usepackage{amsmath}
\usepackage{booktabs}

\usepackage{adjustbox}
\usepackage{multirow}
\usepackage{makecell}
\usepackage{subcaption}

\usepackage{newfloat}
\usepackage{listings}
\DeclareCaptionStyle{ruled}{labelfont=normalfont,labelsep=colon,strut=off} 
\floatstyle{ruled}
\newfloat{listing}{tb}{lst}{}
\floatname{listing}{Listing}
\title{NanoControl: A Lightweight Framework for Precise and Efficient Control in Diffusion Transformer}
\author{
    Shanyuan Liu\textsuperscript{\rm 1 }\thanks{Equal contribution. \textsuperscript{\rm 1}360 AI Research \textsuperscript{\rm 2}Nanjing University of Science and Technology \textsuperscript{\rm 3}University of Science and Technology Beijing \textsuperscript{\rm 4}Beijing University of Aeronautics and Astronautics.
    Corresponding author: Dawei Leng (lengdawei@360.cn).},
    Jian Zhu\textsuperscript{\rm 1 2 * },
    Junda Lu\textsuperscript{\rm 1 3},
    Yue Gong\textsuperscript{\rm 1 4},
    Liuzhuozheng Li\textsuperscript{\rm 1}, \\
    Bo Cheng\textsuperscript{\rm 1},
    Yuhang Ma\textsuperscript{\rm 1},
    Liebucha Wu\textsuperscript{\rm 1},
    Xiaoyu Wu\textsuperscript{\rm 1},
    Dawei Leng\textsuperscript{\rm 1}, 
    Yuhui Yin\textsuperscript{\rm 1},
}
\affiliations{

}

\usepackage{bibentry}

\begin{document}

\maketitle

\begin{abstract}

Diffusion Transformers (DiTs) have demonstrated exceptional capabilities in text-to-image synthesis. However, in the domain of controllable text-to-image generation using DiTs, most existing methods still rely on the ControlNet paradigm originally designed for UNet-based diffusion models. This paradigm introduces significant parameter overhead and increased computational costs.
To address these challenges, we propose the Nano Control Diffusion Transformer (NanoControl), which employs Flux as the backbone network. Our model achieves state-of-the-art controllable text-to-image generation performance while incurring only a 0.024\% increase in parameter count and a 0.029\% increase in GFLOPs, thus enabling highly efficient controllable generation. Specifically, rather than duplicating the DiT backbone for control, we design a LoRA-style (low-rank adaptation) control module that directly learns control signals from raw conditioning inputs. Furthermore, we introduce a KV-Context Augmentation mechanism that integrates condition-specific key-value information into the backbone in a simple yet highly effective manner, facilitating deep fusion of conditional features. Extensive benchmark experiments demonstrate that NanoControl significantly reduces computational overhead compared to conventional control approaches, while maintaining superior generation quality and achieving improved controllability.


\end{abstract}

\begin{figure}[t]  
  \centering
  \includegraphics[width=\linewidth]{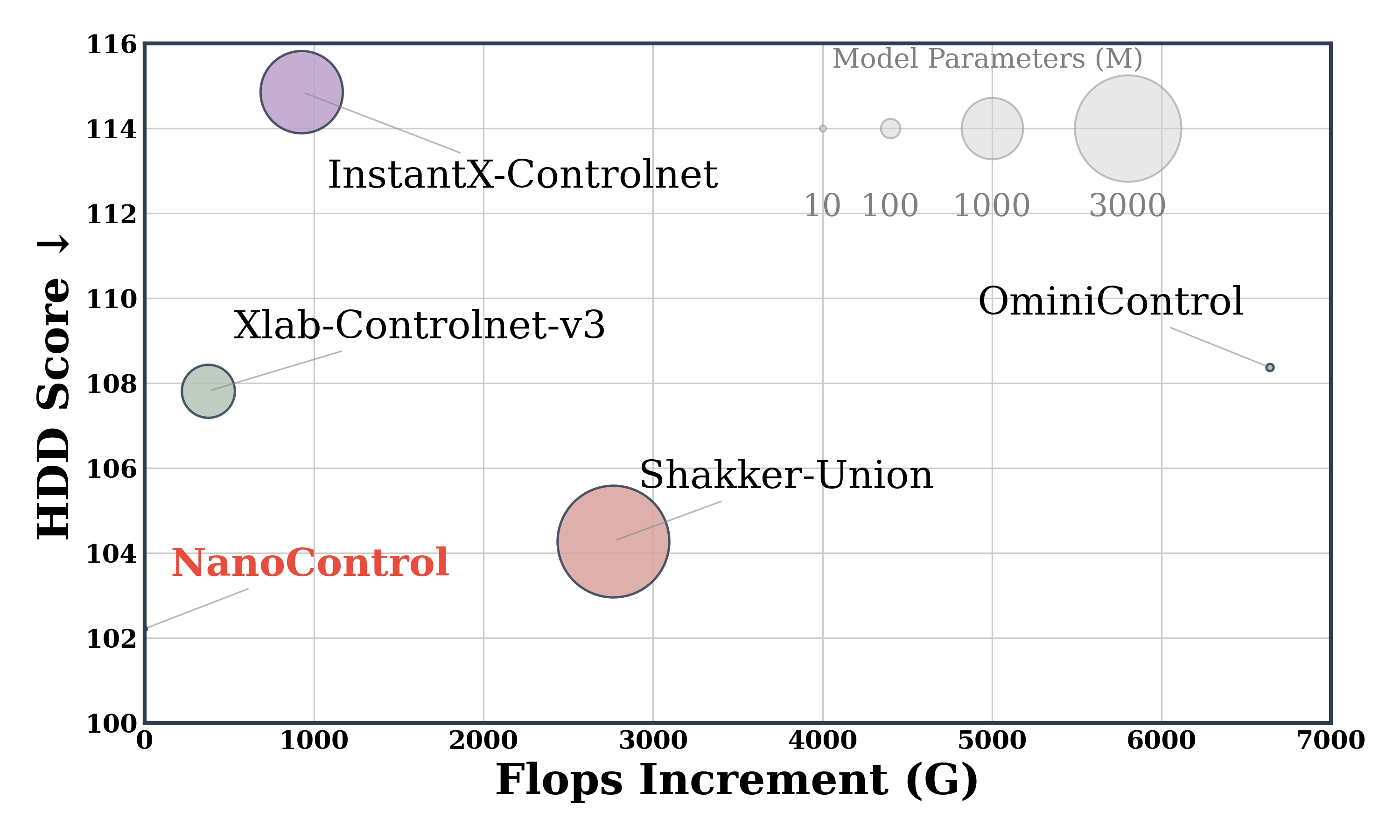}
  \caption{Comparison of HDD Score versus FLOPs increment across different models. Circle size indicates model parameter count (in millions). Our proposed model, NanoControl, achieves competitive performance with the lowest computational cost and moderate parameter size.}
  \label{fig:dit-control}
\end{figure}


\section{Introduction}
Diffusion models have recently achieved state-of-the-art performance in text-to-image generation tasks, with representative models including Stable Diffusion \cite{Rombach2021HighResolutionIS}, DALL·E 2 \cite{Ramesh2022HierarchicalTI},  PixArt-$\alpha$ \cite{Chen2023PixArtFT} and FLUX.1 dev \cite{flux2024}. These models generate high-quality images by leveraging the interaction between textual and visual information. However, relying solely on textual input often fails to fully convey user intent, making it difficult to produce images that accurately match user expectations. To address this limitation, researchers have proposed a range of image-based conditional generation methods \cite{Zhang2023AddingCC,Zhao2023UniControlNetAC,Li2023BLIPDiffusionPS,Li2024ControlARCI}, which incorporate auxiliary inputs—such as Canny edge maps and depth maps—to guide the generation process with greater precision.

\begin{figure*}[t]
    \centering
    \includegraphics[width=\textwidth]{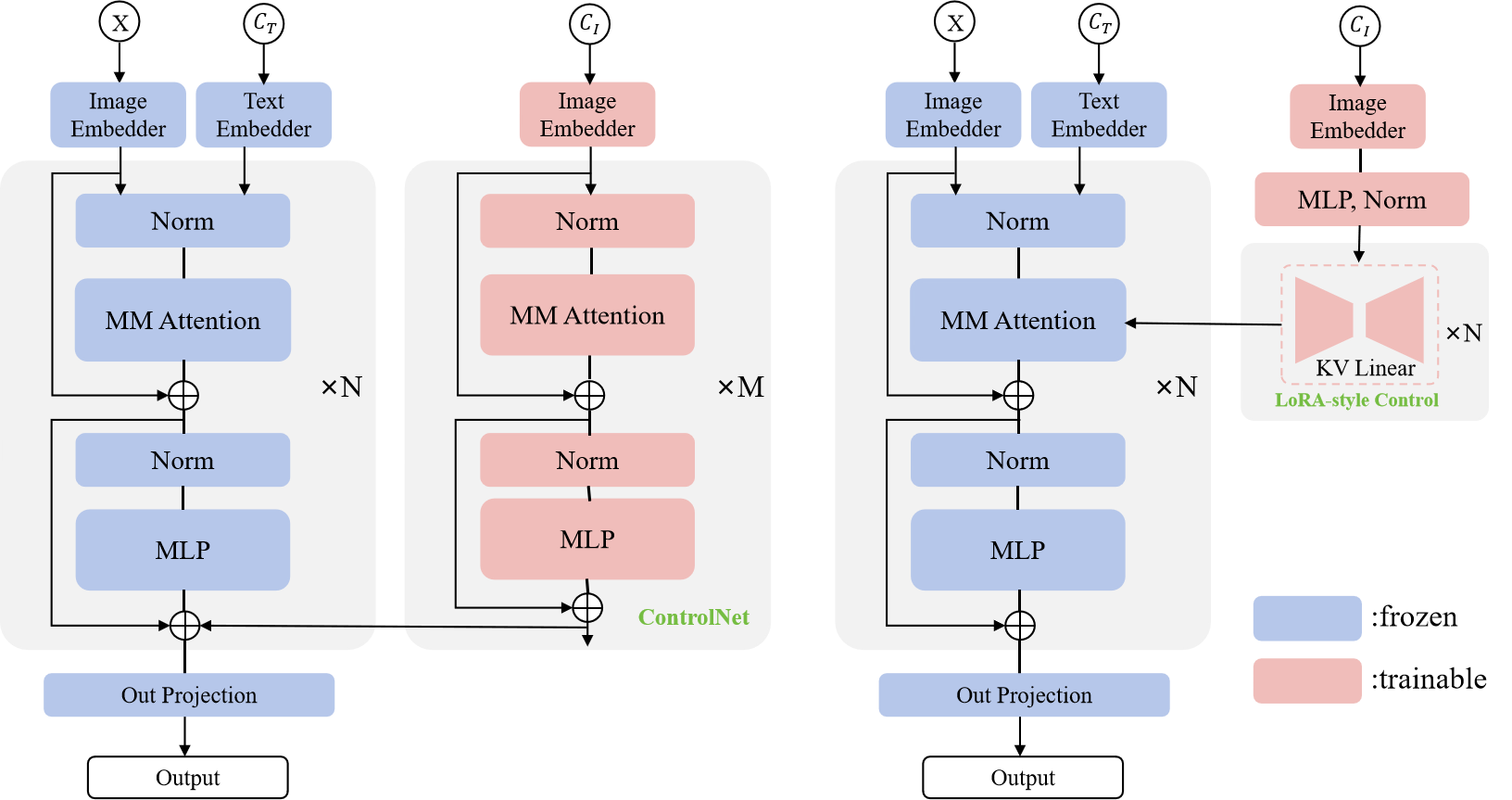}
    \caption{The overall framework of NanoControl and comparison with ControlNet.}
    \label{fig1}
\end{figure*}

Although current methods have achieved significant progress in controllable image generation, the field still faces several challenges, such as limited control precision, large model sizes, and high computational costs. The prevailing approach is to directly adapt control mechanisms originally designed for UNet-based \cite{ronneberger2015u} diffusion models—such as ControlNet \cite{Zhang2023AddingCC}—to the DiT framework \cite{peebles2023scalable, mao2025ace++, chen2024pixart}. The core idea of ControlNet is to enable precise image control by duplicating the backbone network and injecting control conditions, all without compromising the capabilities of the pre-trained model. However, this comes at the cost of a substantial increase in model parameters and computational load, imposing a heavier burden on user devices. An alternative line of work leverages LoRA to enable more efficient training of controllable models with fewer parameters. For instance, OminiControl \cite{Tan2024OminiControlMA} adopts lightweight LoRA fine-tuning along with a unified sequence processing strategy to avoid costly full-parameter updates. Nevertheless, this approach routes all conditional information through the entire backbone, significantly increasing inference-time computational cost (as illustrated in Figure \ref{fig:dit-control}).

To address the aforementioned issues, we propose the Nano Control Diffusion Transformer (NanoControl), which enables the generation of images with the highest quality and the strongest controllability, all while introducing minimal additional parameters and computational cost. As illustrated in Figure \ref{fig:dit-control}, NanoControl achieves state-of-the-art generation quality while reducing the additional parameter count and FLOPs by several orders of magnitude compared to existing methods. Specifically, We introduce two LoRA-style Control Module branches alongside the attention layers of each MM-DiT block, which directly process the input conditioning signals to generate condition-specific key and value representations. Then, in each attention layer, we introduce a KV-Context Augmentation Mechanism, which combines the original key and value sequences from the backbone with the condition-specific counterparts, thereby enabling a direct and efficient injection of control information.

Benefited from our design, the control-related parameter overhead is inherently minimal—almost negligible in practice. In terms of computation, unlike previous LoRA-based control models that route the conditioning information through the full backbone, our approach computes it directly using a lightweight LoRA-style Control Module, resulting in extremely low additional computational cost. Moreover, this design enables more direct and efficient utilization of conditioning signals, which significantly enhances the controllability of the generated images.

In summary, our main contributions are as follows:

\begin{itemize}
    \item We propose NanoControl (Nano Control Diffusion Transformer), which achieves highly effective conditional text-to-image generation with only 0.024\% additional parameters and 0.029\% extra FLOPs, making it one of the most lightweight control mechanisms to date.
    \item We design a LoRA-style independent control branch to extract condition features, and introduce the KV-Context Augmentation Mechanism to inject them into the backbone effectively, leading to improved control fidelity.
    \item Extensive experiments demonstrate that our method achieves state-of-the-art controllability while maintaining extremely low parameter and computation overhead.
\end{itemize}

\begin{table*}[t]
    \centering
    \footnotesize
    \begin{adjustbox}{width=\textwidth}
    \begin{tabular}{ccccccccc}
        \toprule
        \multirow{2}{*}{Task} & \multirow{2}{*}{Methods}
        & \multicolumn{1}{c}{Controllability} & \multicolumn{2}{c}{Image Quality} & \multicolumn{2}{c}{Consistency} \\
        \cmidrule(lr){3-3} \cmidrule(lr){4-5} \cmidrule(lr){6-7}
        & & HDD$\downarrow$ / MSE$\downarrow$ & FID$\downarrow$ & MUSIQ$\uparrow$  & CLIP Text$\uparrow$ & CLIP Image$\uparrow$ \\
        \midrule
        \multirow{6}{*}{Canny} 
        & Flux-Contolnet      & 104.48  & 20.04   & \underline{71.85}  & 0.238   & 0.704 \\
        & InstantX-Controlnet & 114.86  & 30.36   & 52.31  & 0.234   & 0.701 \\
        & Shakker-Union       & \underline{104.28}  & 41.82   & 63.10  & 0.231   & 0.699 \\
        & Xlab-Controlnet-v3  & 107.81  & 24.63   & 68.90  & \textbf{0.254}   & 0.706 \\
        & OminiControl        & 108.37  & \textbf{16.74}   & 70.91  & 0.252   & \underline{0.790} \\
        & Ours                & \textbf{102.22}  & \underline{16.99}   & \textbf{71.86}  & \textbf{0.254}   & \textbf{0.793} \\
        \midrule
        \multirow{6}{*}{Depth} 
        & Flux-Contolnet      & \underline{628.74}  & \textbf{15.35}   & 70.66  & 0.249   & 0.741 \\
        & InstantX-Controlnet & 1144.19 & 36.25   & 61.61  & 0.228   & 0.678 \\
        & Shakker-Union       & 1010.42 & 24.95   & 67.41  & 0.241   & 0.712 \\
        & Xlab-Controlnet-v3  & 4205.97 & 23.69   & 70.44  & \textbf{0.259}   & 0.714 \\
        & OminiControl        & 719.60  & \underline{19.00}   & \underline{71.06}  & 0.250   & \underline{0.746} \\
        & Ours                & \textbf{567.95}  & 19.48   & \textbf{71.92}  & \underline{0.253}   & \textbf{0.759} \\
        \midrule
        \multirow{4}{*}{Colorization} 
        & InstantX-Controlnet       & 662.21  & 17.95   & 57.51  & 0.248   & 0.873 \\
        & Shakker-Union    & 693.56  & 17.95   & 55.43  & \textbf{0.251}   & 0.872 \\
        & OminiControl & \textbf{126.31}  & \textbf{10.17}   & \textbf{67.63}  & \underline{0.250}   & \textbf{0.911} \\
        & Ours                & \underline{129.57}  & \underline{10.95}   & \underline{67.07}  & 0.249   & \textbf{0.911} \\
        \midrule
        \multirow{2}{*}{Hed} 
        & Xlab-Controlnet-v3  & 124.82  & 26.29   & 68.83  & \textbf{0.255}   & 0.707 \\
        & Ours                & \textbf{120.57}  & \textbf{16.07}   & \textbf{70.60}  & 0.252   & \textbf{0.819} \\
        \midrule
        \addlinespace[0.01em] 
        \midrule
        \multirow{6}{*}{\makecell{Canny*}} 
        & Flux-Contolnet            & 120.51  & 78.02   & \underline{68.62}     & 0.244     & 0.767 \\
        & InstantX-Controlnet       & 124.95  & 79.96   & 49.84     & 0.239     & \underline{0.786} \\
        & Shakker-Union             & 114.96  & 81.41   & 55.47     & 0.239     & 0.782 \\
        & Xlab-Controlnet-v3        & \textbf{109.68}  & 84.17   & 59.09     & \underline{0.253}     & 0.750 \\
        & OminiControl              & 135.86  & \underline{77.15}   & 63.84     & 0.249     & 0.783 \\
        & Ours                      & \underline{114.03}  & \textbf{73.63}   & \textbf{69.26}     & \textbf{0.255}     & \textbf{0.816} \\
        \bottomrule
    \end{tabular}
    \end{adjustbox}
    \caption{We evaluate the baseline methods and NanoControl in terms of controllability (using HDD for Canny and MSE for the others), image quality (FID and MUSIQ), and consistency (CLIP-Text and CLIP-Image). Best results are shown in \textbf{bold}, and second-best results are \underline{underlined}. Quantitative evaluation of Canny* was conducted on 500 images at a resolution of 1024×1024.}
    \label{tab:compare}
\end{table*}

\section{Related works}
\label{related_works}

\subsection{Diffusion models}
Diffusion models \cite{ho2020denoising, song2020score} have experienced rapid progress in recent years, especially in the field of image generation. Text-to-image synthesis has seen particularly significant advances, notably following the introduction of latent diffusion models \cite{rombach2022high, podell2023sdxl, saharia2022photorealistic}. These methods utilize text embeddings—extracted from large pre-trained language models \cite{Radford2021LearningTV, raffel2020exploring}—and incorporate them into the latent space via cross-attention mechanisms, enabling the generation of high-fidelity and diverse images conditioned on textual input. To further enhance generative performance, transformer architectures \cite{vaswani2017attention} have recently been adopted within diffusion frameworks. Models such as DiT \cite{peebles2023scalable} exemplify this trend, leveraging the power of transformers to model global image dependencies through denoising processes, leading to superior image quality. Building upon the DiT architecture, FLUX \cite{flux2024} integrates the concept of flow matching, which aligns the generative trajectory with the data distribution more effectively, achieving even more compelling and robust results.

\subsection{Controllable generation}
While conditioning on text alone has driven remarkable advances in image generation, it often fails to fully capture users’ intentions. Ambiguous descriptions, incomplete specifications, or the lack of precise vocabulary can result in generated images that significantly deviate from expectations. To overcome these limitations, recent approaches have explored incorporating additional control signals—particularly visual inputs—to enhance controllability. A prominent example is ControlNet \cite{Zhang2023AddingCC}, which introduces learnable control modules that integrate auxiliary inputs (e.g., edge maps, depth maps) into pre-trained diffusion models, enabling precise and fine-grained control over image structure and content. T2I-Adapter \cite{Mou2023T2IAdapterLA} employs lightweight adapters to support multimodal conditioning, while IP-Adapter \cite{Ye2023IPAdapterTC} adopts a decoupled cross-attention mechanism to facilitate image-based prompting. However, these methods are primarily designed for UNet-based Stable Diffusion \cite{Rombach2021HighResolutionIS} models and are not readily transferable to transformer-based architectures like DiT \cite{Peebles2022ScalableDM}. Addressing this gap, OminiControl \cite{Tan2024OminiControlMA} proposes a unified sequence processing framework with LoRA \cite{Hu2021LoRALA} fine-tuning, offering a minimal yet generalizable control mechanism for DiT models. Despite these developments, there remains a strong demand for control frameworks that simultaneously achieve high controllability, computational efficiency, and superior image quality.

\section{Methods}
\label{methods}

\subsection{Preliminary}
\subsubsection{Diffusion Transformer}
Currently, state-of-the-art open-source text-to-image models such as Flux \cite{flux2024}, Stable Diffusion 3 \cite{Rombach2021HighResolutionIS}, and PixArt-$\alpha$ \cite{Chen2023PixArtFT} are all built upon the DIT architecture. DIT has also demonstrated remarkable performance in text-to-video generation. It employs multiple layers of transformer blocks as a denoising network, iteratively refining noisy image representations to produce high-quality outputs. A prominent example is MMDIT in FLUX, which processes two types of tokens: noisy image tokens $X \in \mathbb{R}^{N \times d}$ and text condition tokens $C_T \in \mathbb{R}^{M \times d}$, where $d$ denotes the embedding dimension, and $N$ and $M$ represent the number of image and text tokens, respectively. The information from these two modalities undergoes multimodal fusion within the network while maintaining a constant token count throughout the process.

\subsubsection{Controllable Diffusion Models}
A prevalent strategy for enhancing diffusion models involves incorporating spatial control conditions into the base model, thereby extending their capabilities beyond text-based guidance. For example, in diffusion models utilizing the UNet architecture, ControlNet \cite{Zhang2023AddingCC} employs a parallel side branch initialized with a copy of the main network's parameters. A zero convolution layer is introduced at the merging stage to ensure a stable training initialization. This approach enables the integration of spatial conditions such as Canny edges, depth maps, and sketches into the model, empowering users to generate images that align more closely with their intended specifications.

\begin{equation}
\boldsymbol{y}_{\text{c}} = \mathcal{F}(\boldsymbol{x} ; \Theta) + \mathcal{Z}_2\left( \mathcal{F}c\left( \boldsymbol{x} + \mathcal{Z}1(\boldsymbol{c} ; \Theta{z1}) ; \Theta_c \right) ; \Theta{z2} \right)
\end{equation}

Herein, \(\mathcal{F}(\boldsymbol{x} ; \Theta)\) is the frozen original model (e.g., the U-Net block of Stable Diffusion), with parameters \(\Theta\) kept fixed. \(\mathcal{F}_c(\cdot ; \Theta_c)\) is the trainable copy model, whose parameters \(\Theta_c\) are learnable. \(\mathcal{Z}_1\) and \(\mathcal{Z}_2\) are zero convolution layers (1×1 convolutions initialized with zero weights), with parameters \(\Theta_{z1}\) and \(\Theta_{z2}\) updated during training. \(\boldsymbol{c}\) denotes the conditional input (e.g., edge maps, pose maps), which is transformed into the feature space by \(\mathcal{Z}_1\) and then added to the original input \(\boldsymbol{x}\).


\subsection{NanoControl}

\begin{figure*}[t]
    \centering
    \setlength{\tabcolsep}{1pt} 

    \begin{tabular}{cccccccc}
        \scriptsize Condition & 
        \scriptsize Original & 
        \scriptsize Flux & 
        \scriptsize InstantX & 
        \scriptsize Shakker & 
        \scriptsize Xlab & 
        \scriptsize OminiControl &
        \scriptsize \textbf{NanoControl} \\

        \includegraphics[width=0.12\textwidth]{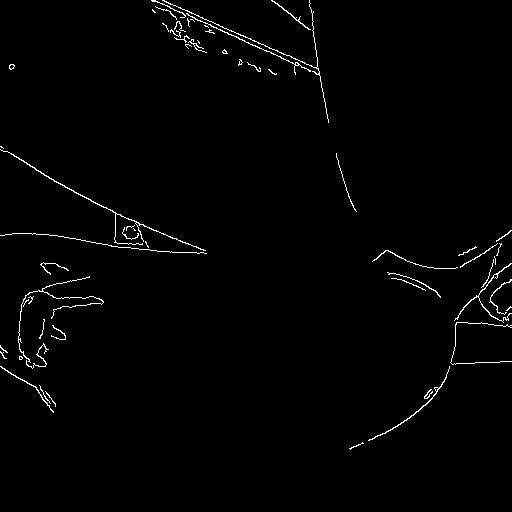} &
        \includegraphics[width=0.12\textwidth]{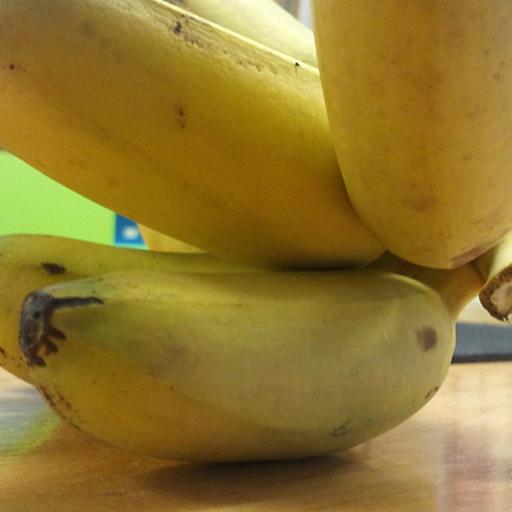} &
        \includegraphics[width=0.12\textwidth]{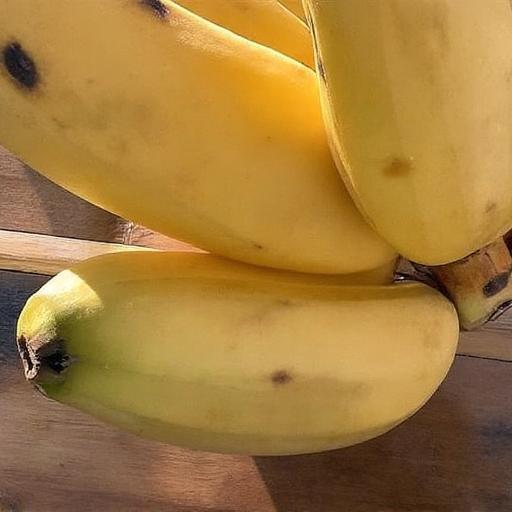} &
        \includegraphics[width=0.12\textwidth]{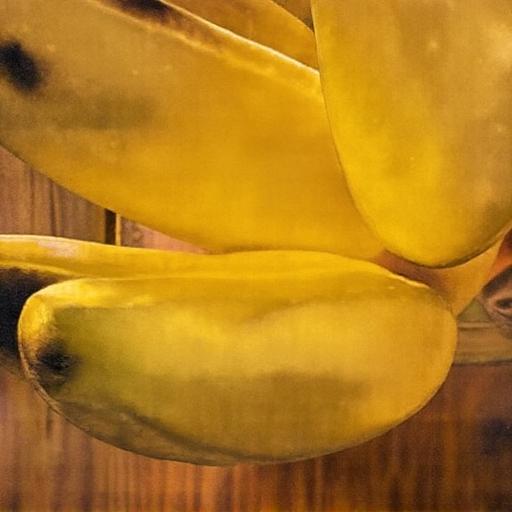} &
        \includegraphics[width=0.12\textwidth]{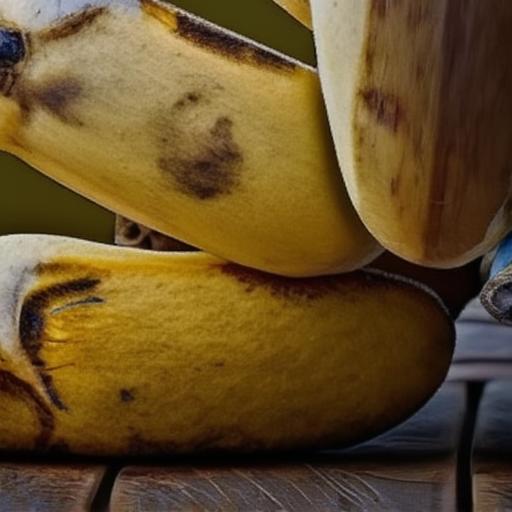} &
        \includegraphics[width=0.12\textwidth]{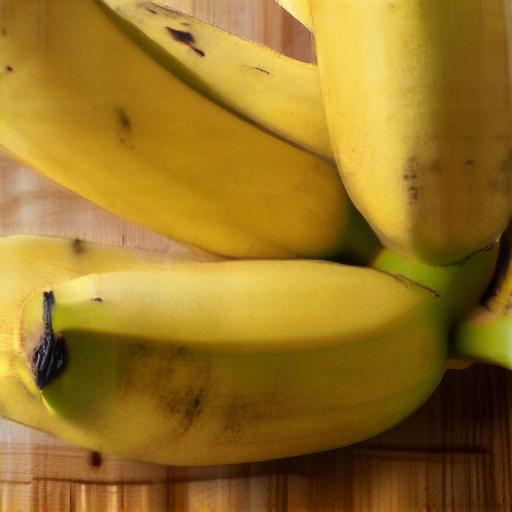} &
        \includegraphics[width=0.12\textwidth]{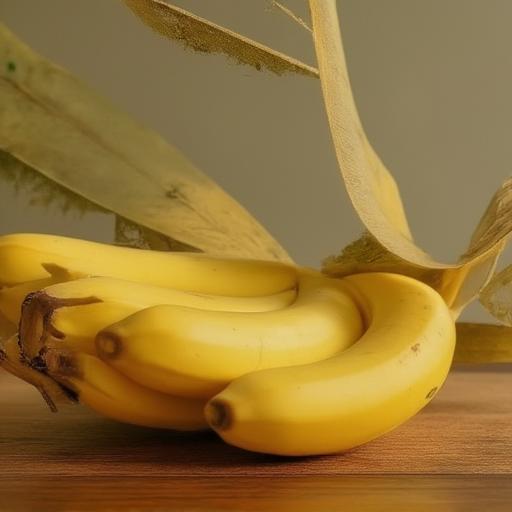} &
        \includegraphics[width=0.12\textwidth]{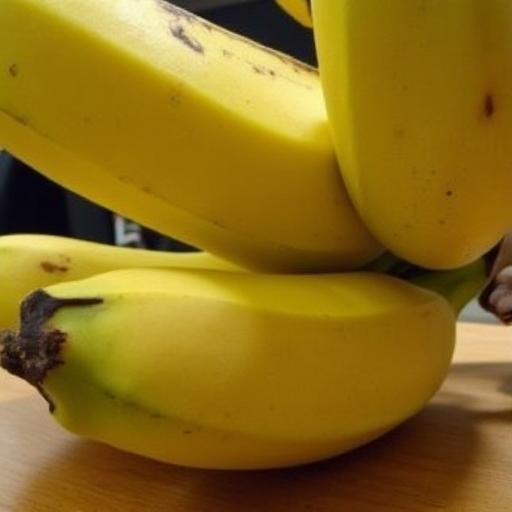} \\
        \multicolumn{8}{l}{\scriptsize \textbf{Canny:} A bunch of bananas sitting on top of a wooden table.} \\

        \includegraphics[width=0.12\textwidth]{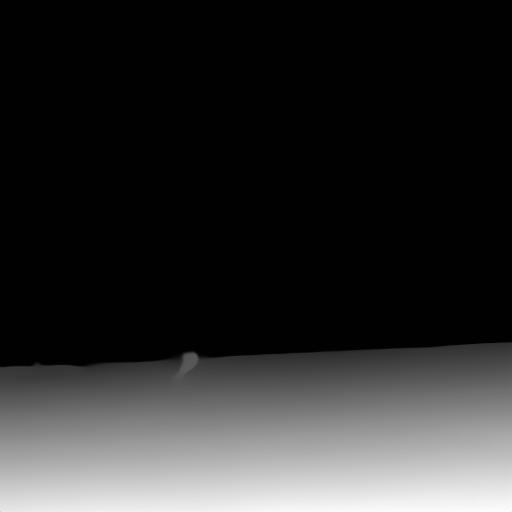} &
        \includegraphics[width=0.12\textwidth]{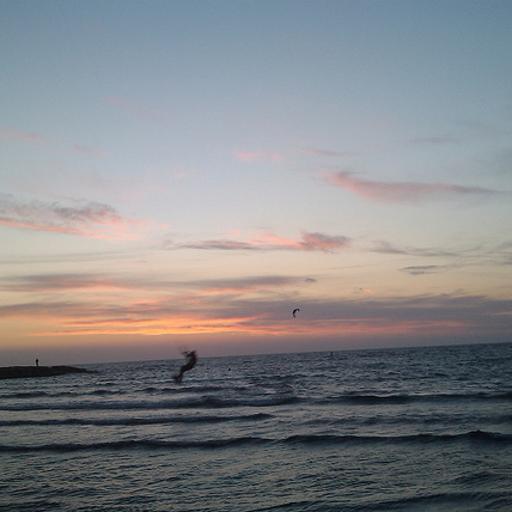} &
        \includegraphics[width=0.12\textwidth]{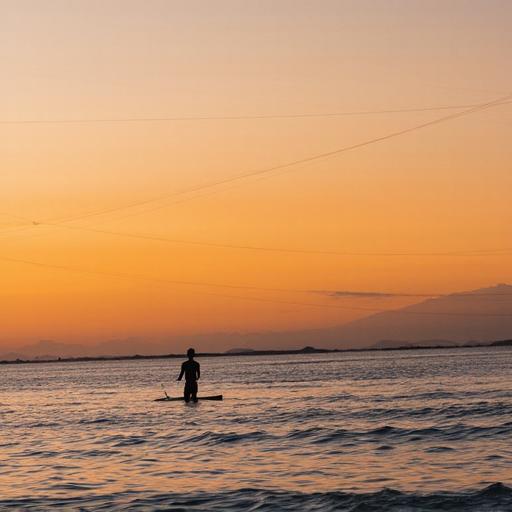} &
        \includegraphics[width=0.12\textwidth]{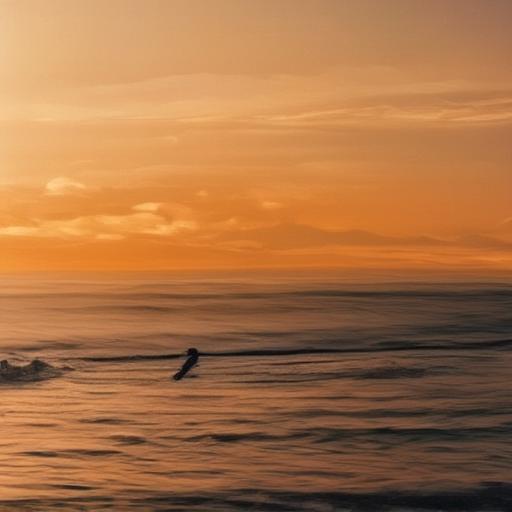} &
        \includegraphics[width=0.12\textwidth]{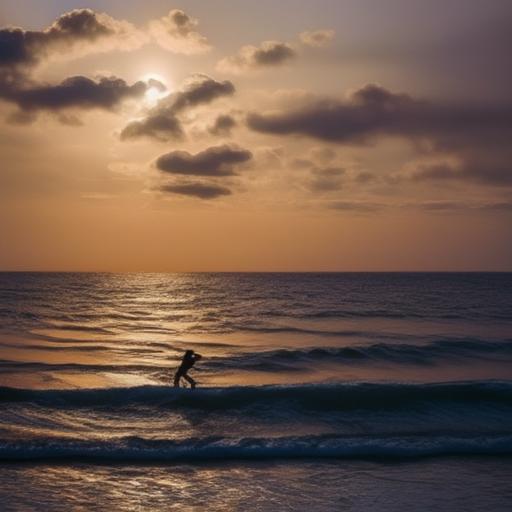} &
        \includegraphics[width=0.12\textwidth]{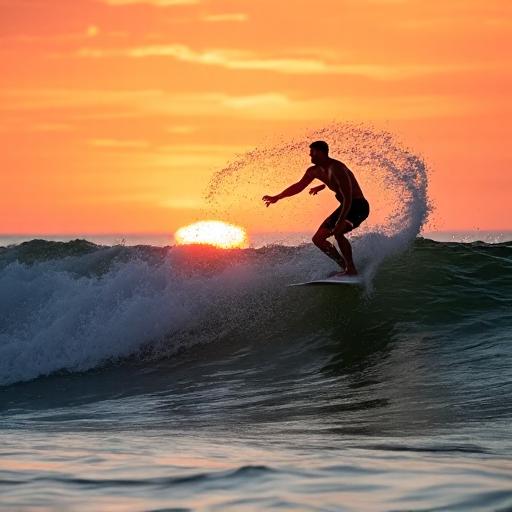} &
        \includegraphics[width=0.12\textwidth]{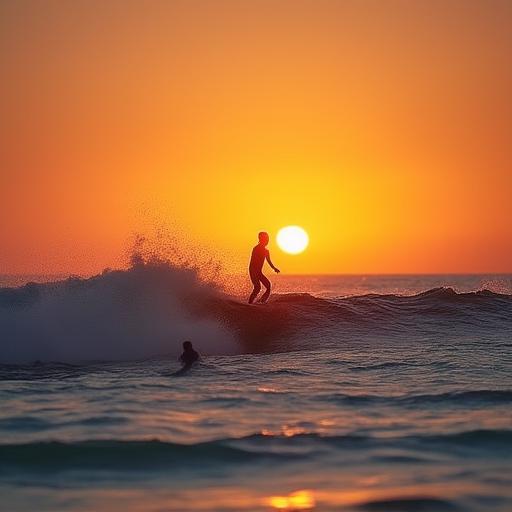} &
        \includegraphics[width=0.12\textwidth]{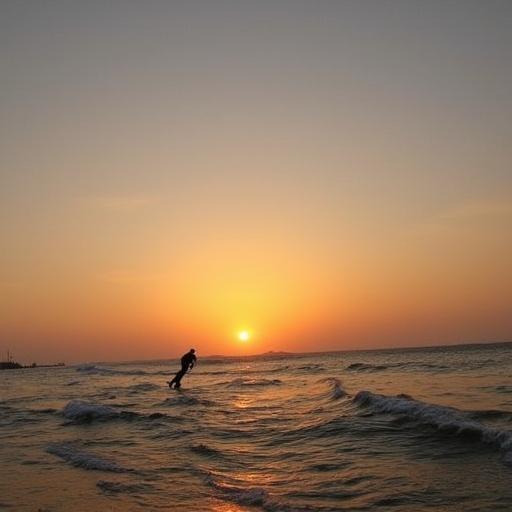} \\
        \multicolumn{8}{l}{\scriptsize \textbf{Depth:} A man surfing in the ocean as the sun sets.} \\

        \includegraphics[width=0.12\textwidth]{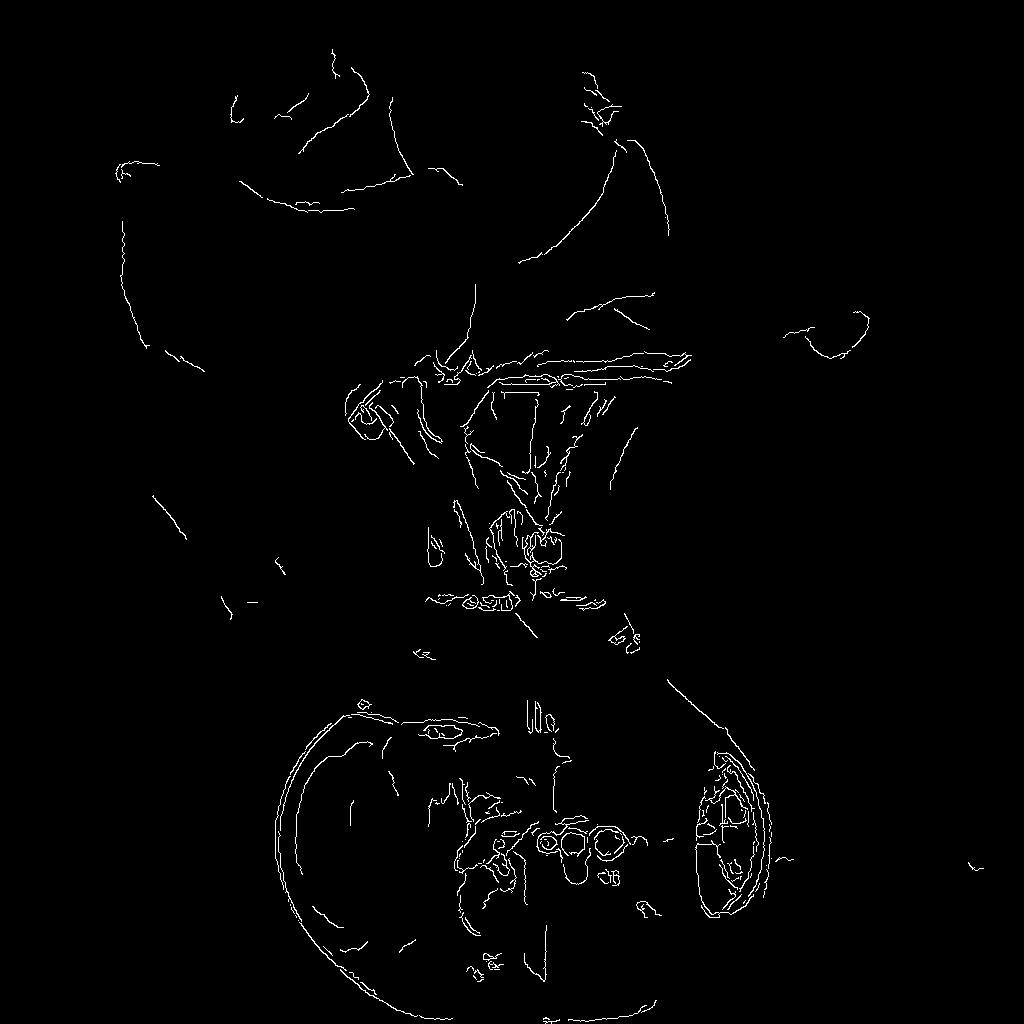} &
        \includegraphics[width=0.12\textwidth]{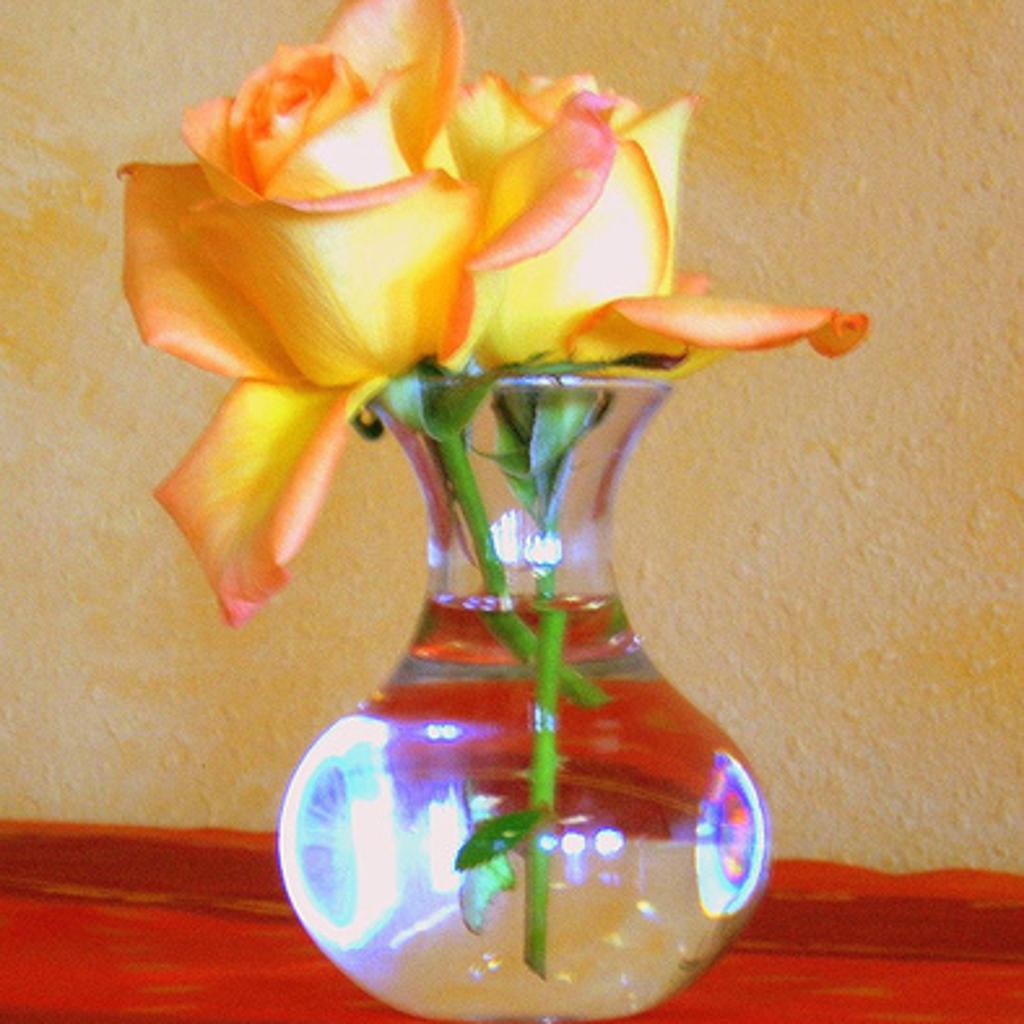} &
        \includegraphics[width=0.12\textwidth]{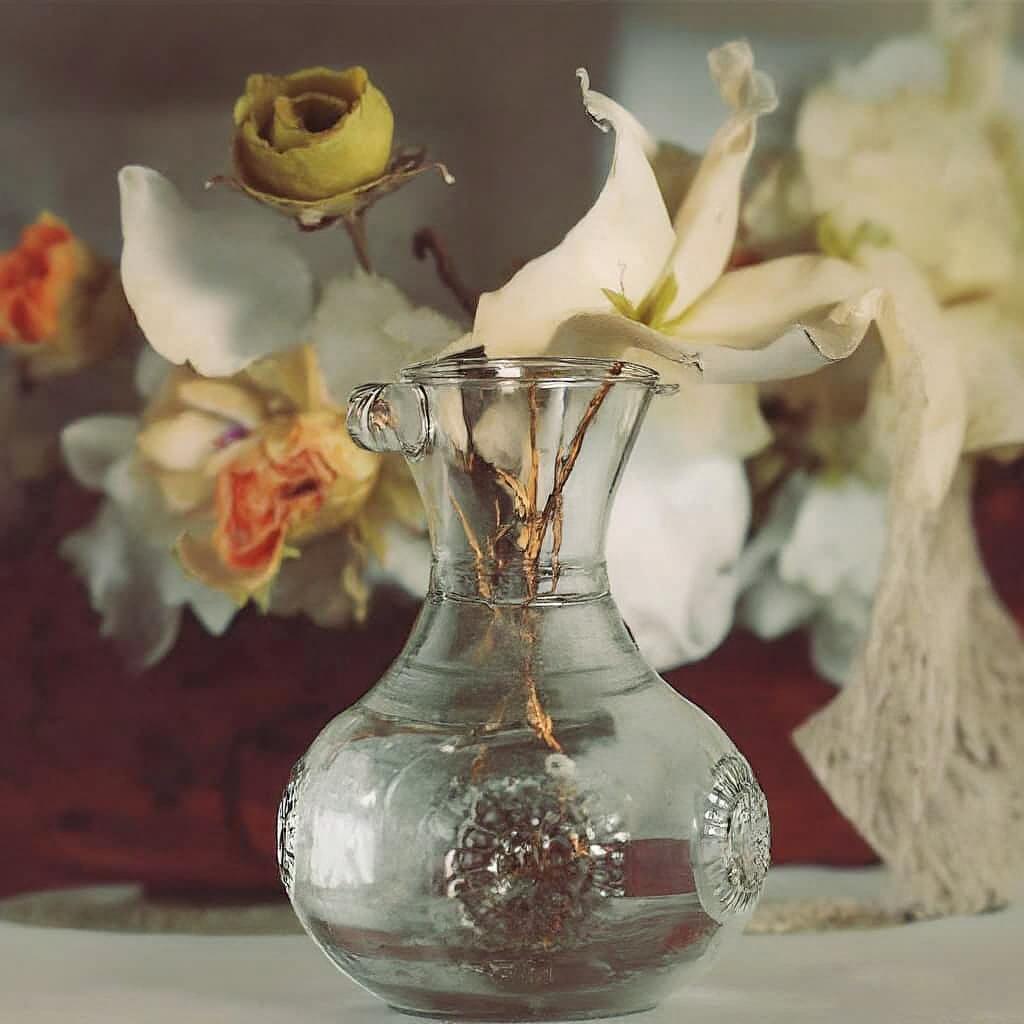} &
        \includegraphics[width=0.12\textwidth]{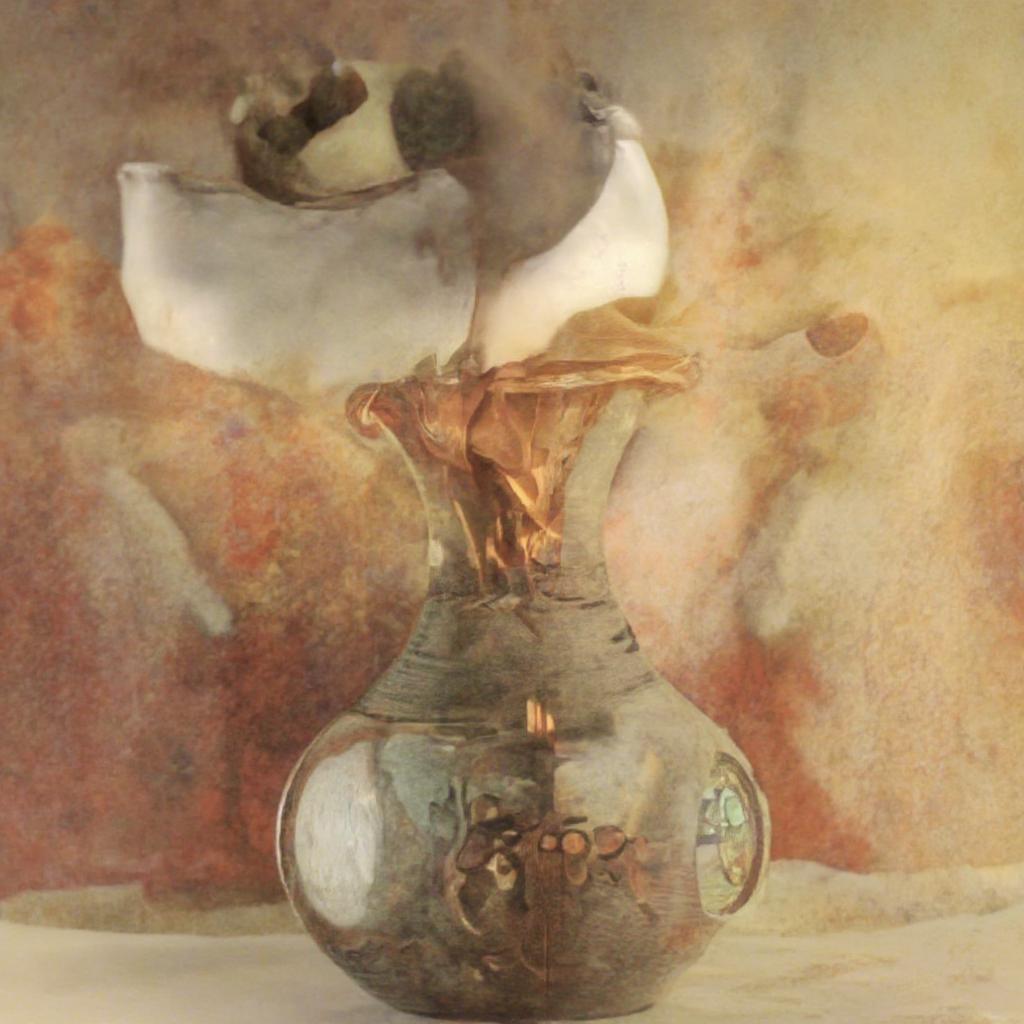} &
        \includegraphics[width=0.12\textwidth]{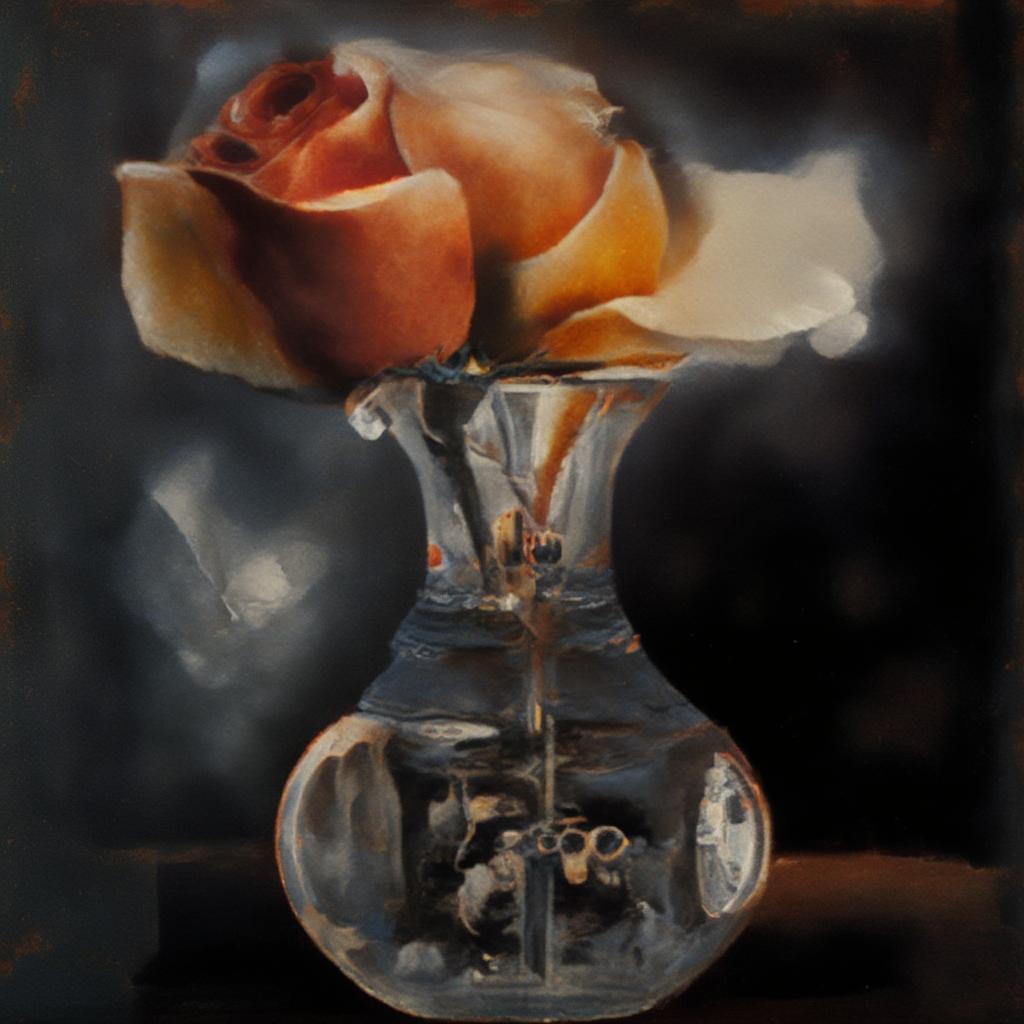} &
        \includegraphics[width=0.12\textwidth]{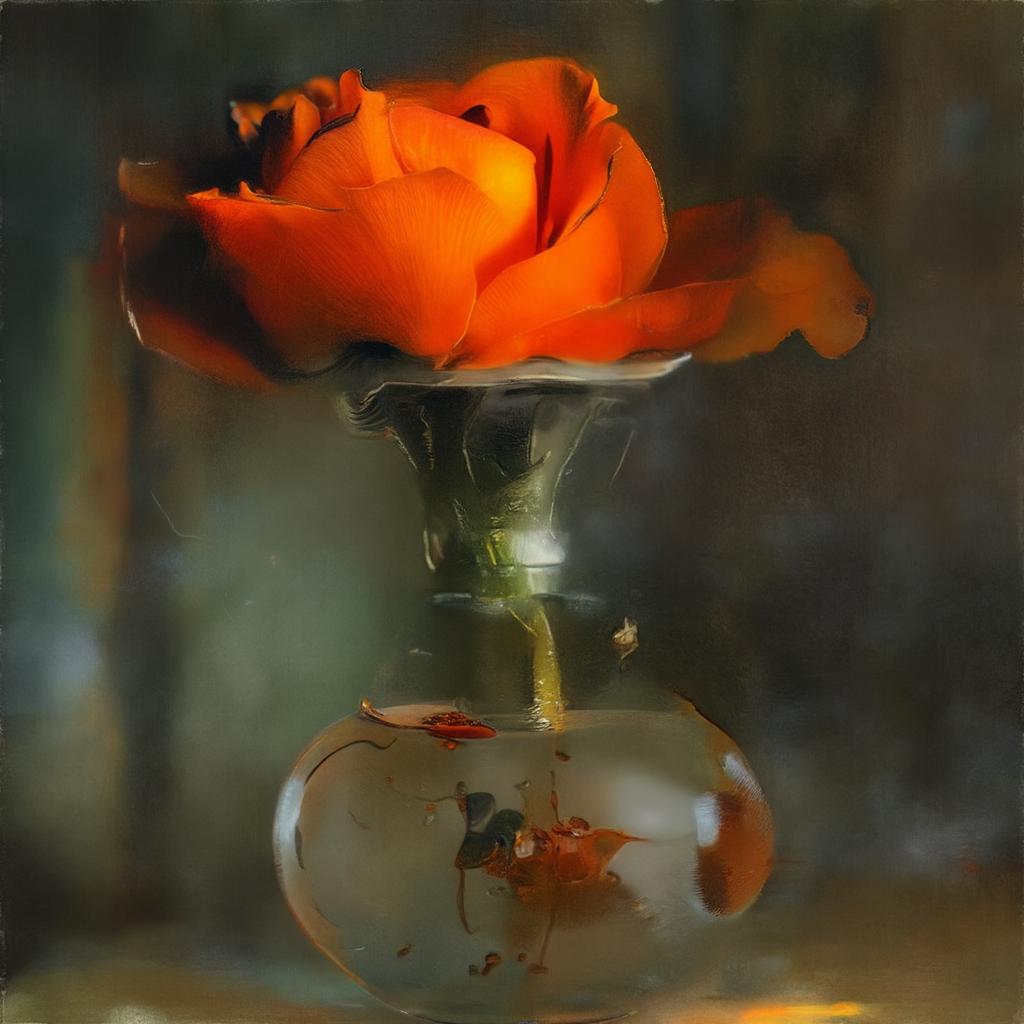} &
        \includegraphics[width=0.12\textwidth]{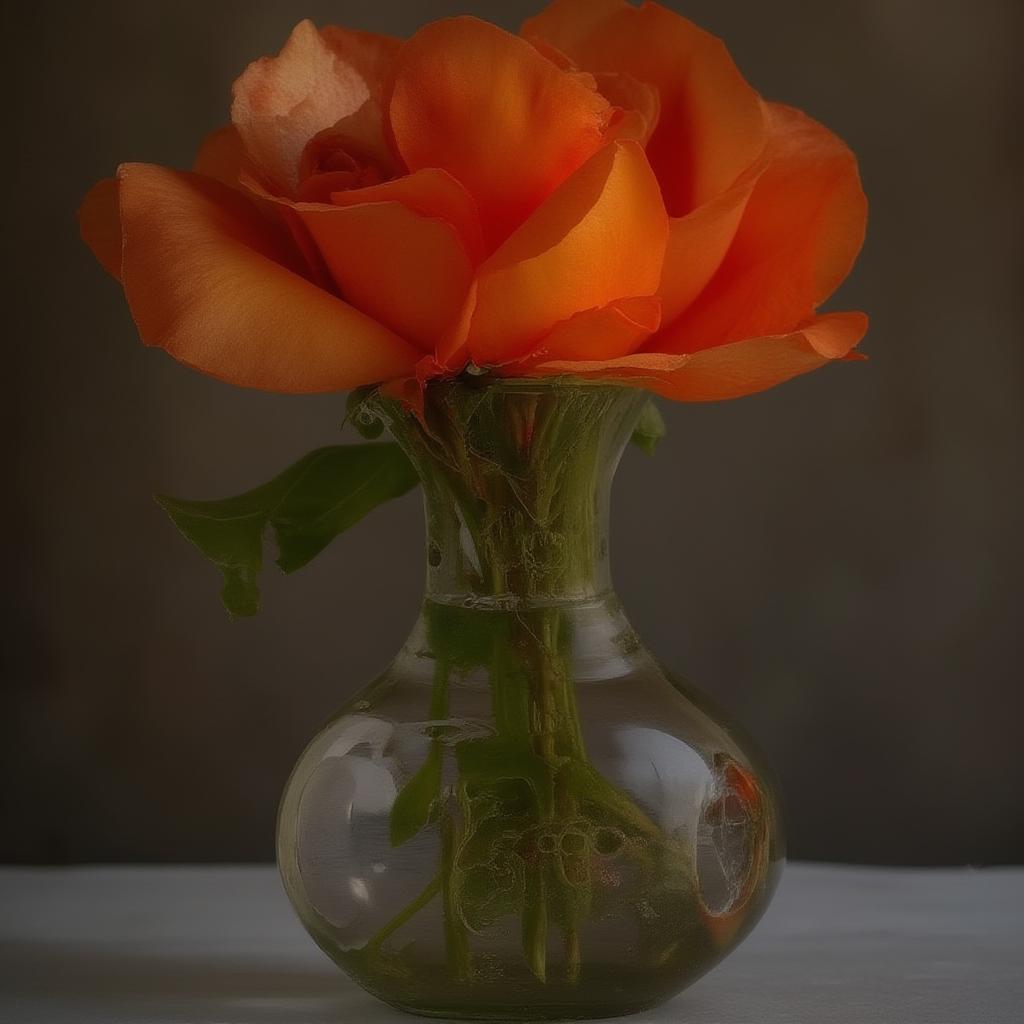} &
        \includegraphics[width=0.12\textwidth]{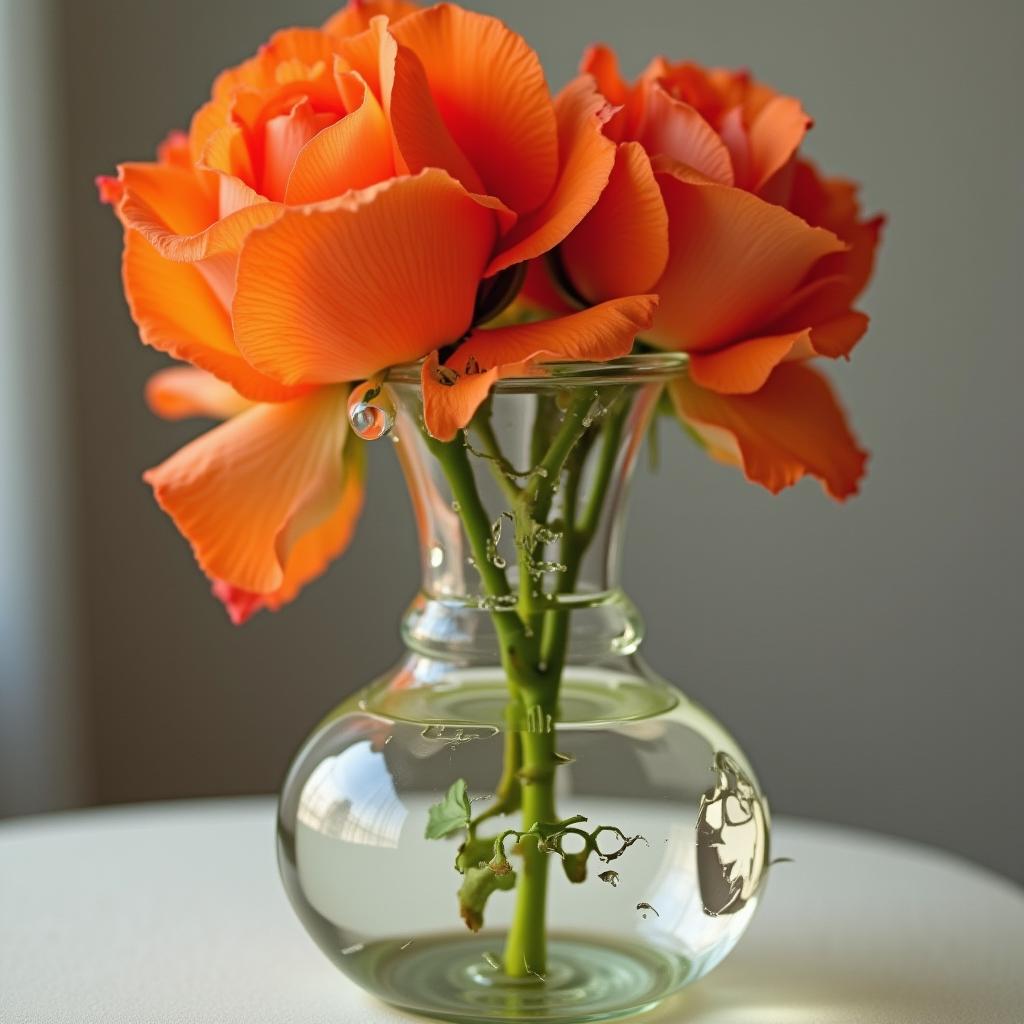} \\
        \multicolumn{8}{l}{\scriptsize \textbf{Canny 1024:} An orange reddish rose in a vase filled with water on top of a table.} \\

        \scriptsize Condition & 
        \scriptsize InstantX & 
        \scriptsize Shakker & 
        \scriptsize OminiControl & 
        \scriptsize \textbf{NanoControl} & 
        \scriptsize Condition & 
        \scriptsize Xlab &
        \scriptsize \textbf{NanoControl} \\

        \includegraphics[width=0.12\textwidth]{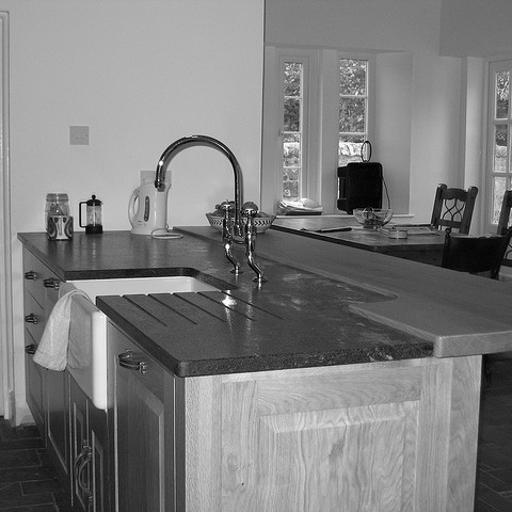} &
        \includegraphics[width=0.12\textwidth]{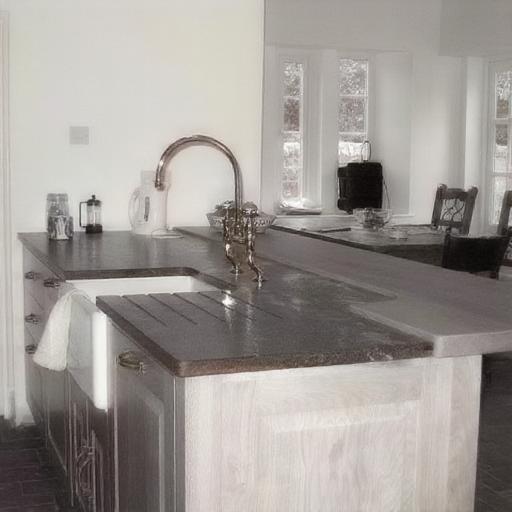} &
        \includegraphics[width=0.12\textwidth]{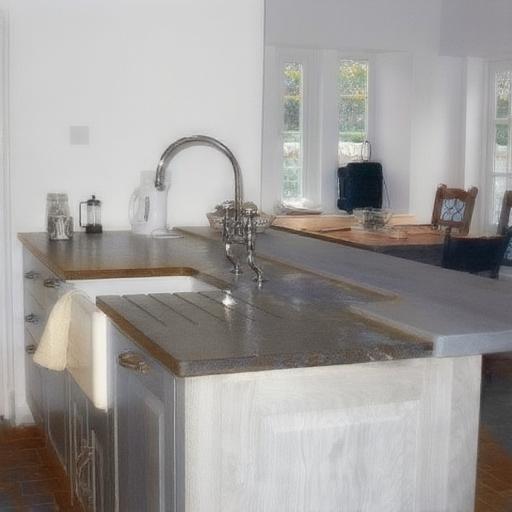} &
        \includegraphics[width=0.12\textwidth]{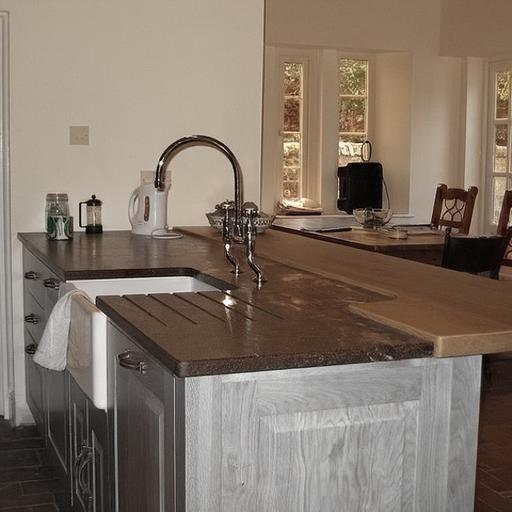} &
        \includegraphics[width=0.12\textwidth]{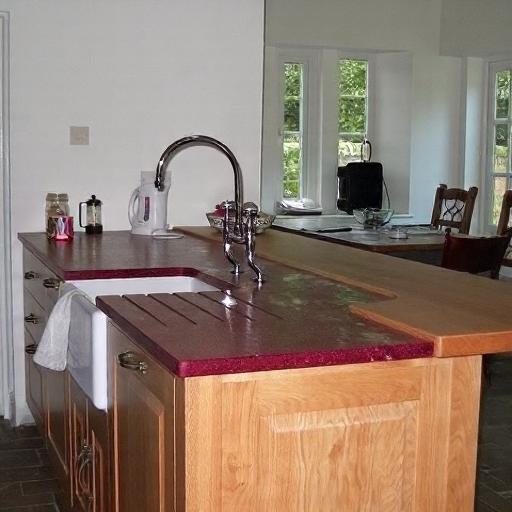} &
        \includegraphics[width=0.12\textwidth]{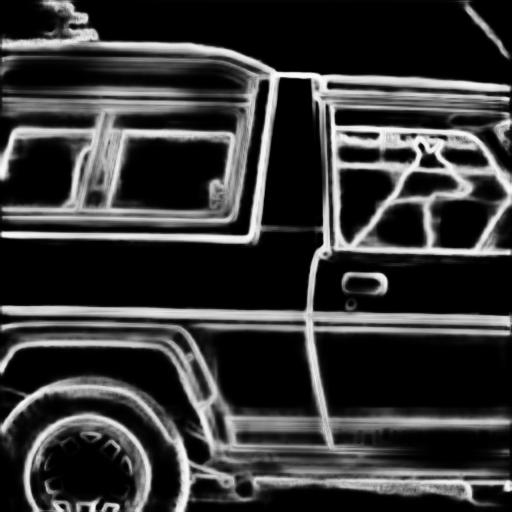} &
        \includegraphics[width=0.12\textwidth]{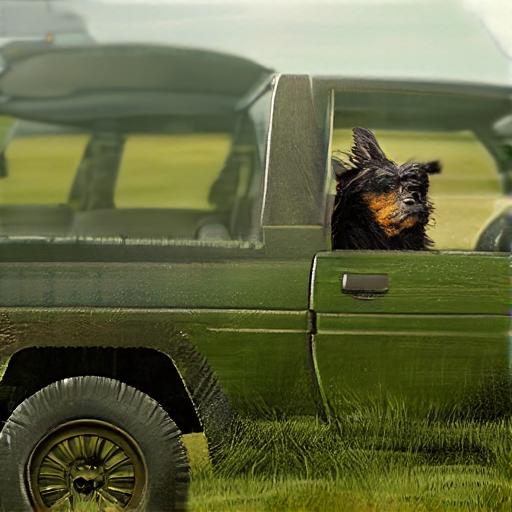} &
        \includegraphics[width=0.12\textwidth]{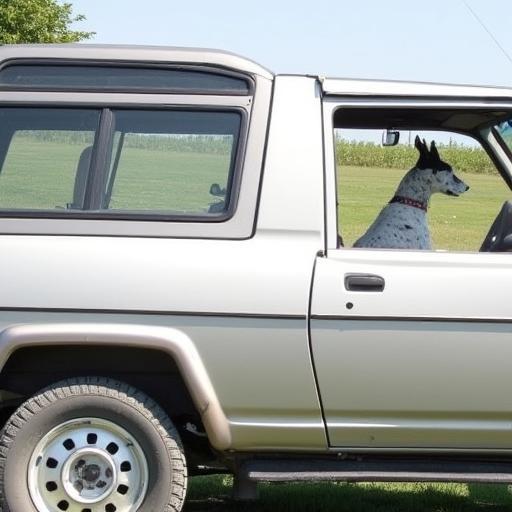} \\
        \multicolumn{5}{l}{\scriptsize \textbf{Color:} A kitchen with a counter and a table with chairs.} &
        \multicolumn{3}{l}{\scriptsize \parbox[t]{0.35\textwidth}{\textbf{HED:} A dog driving an SUV in an open grass covered field.}} \\
        
    \end{tabular}
    \caption{Qualitative comparison with other models under four tasks}
    \label{fig:multi-subfigures}
\end{figure*}

To implement a lightweight efficient control scheme within the DIT architecture and capitalize on the multimodal interaction strengths of MMDIT block, we propose the Nano Control Diffusion Transformer (NanoControl). We conduct experiments using the current state-of-the-art open-source DIT text-to-image model, Flux.1 dev \cite{flux2024}.

\subsubsection{Model}
First, we map the conditional control image to the latent space via the VAE \cite{Lopez2020AUTOENCODINGVB} encoder module of Flux.1 dev. This approach offers two key benefits: 1) The VAE mapping is nearly lossless, reducing token length without sacrificing information; 2) It reuses the original parameters, avoiding additional parameter increases and ensuring compatibility. Then, the conditional features are aligned in feature dimension to be the same as the main backbone features through the learnable Image embedder module shown in Figure \ref{fig1}. After passing through a nonlinear MLP and Norm layers, we obtain the original conditional information \(C_h\). 

Subsequently, we design an independent side branch similar to ControlNet \cite{Zhang2023AddingCC}. This side branch processes conditional information separately and integrates it into the main backbone. The advantage of this design is that it does not affect the main backbone’s parameters, facilitating integration with other backbone plugins such as LoRA \cite{Hu2021LoRALA}. However, the side branch of ControlNet is overly redundant. As stated in several Unet-related papers (\cite{Peng2024ControlNeXtPA},\cite{Zhao2023UniControlNetAC}), various methods have been proposed to reduce the side branch’s parameter count. Unlike ControlNet, to minimize the parameter count of the control module, our independent side branch only contains linear K and V projection layers matching the number of block layers. However, since the feature dimensions in the network are already large—for example, Flux uses a feature dimension of 3072—each linear K and V layer in our independent side branch is a 3072×3072 matrix, which remains excessively large. Therefore, we employ a LoRA-style low-rank adaptation for the K and V matrices, resulting in an ultra-lightweight variant of the NanoControl model, with only a 0.024\% increase in parameters and a 0.029\% increase in GFLOPs in Flux.

Then, in each block, this identical \(C_h\) is input to the K and V projections of the independent side branch for computation. The control function is realized through interaction with the main backbone information via the KV-Context Augmentation Mechanism.

\subsubsection{KV-Context Augmentation Mechanism}

In ControlNet \cite{Zhang2023AddingCC}, conditional information is only injected into the first block, and each subsequent block uses the conditional features output by the previous block. The final processed conditional features are then fused with the corresponding features of the main backbone block via element-wise addition. OminiControl \cite{Tan2024OminiControlMA} follows a similar strategy, injecting the original conditional information only at the first block. Although residual connections mitigate feature vanishing, the original information is still prone to being forgotten.

Unlike the above methods, we do not inject the original conditional features solely into the first block, nor do we rely on simple addition for feature fusion. Instead, we propose a KV-Context Augmentation Mechanism. Specifically, we inject the original conditional information \(C_h\) into each block, meaning that each block does not depend on the conditional feature results of the previous block. Then, in the MMDIT section, the keys (K) and values (V) are concatenated with the main backbone’s text tokens, image tokens, and the conditional tokens processed by the side branch, while the query (Q) still uses the concatenation of the main backbone’s text tokens and image tokens. This allows conditional information to be integrated into the main backbone through a multimodal attention mechanism.

This is our KV-Context Augmentation Mechanism, which ensures that conditional features are not forgotten and enables feature interaction through attention rather than simple addition, fully leveraging the capabilities of the DIT architecture.

\section{Experiment}
\label{experiment}

\subsection{Experimental Setup}

\paragraph{Evaluation Metrics.}
We conduct a comprehensive evaluation of image generation quality from multiple perspectives using a range of metrics. FID \cite{Heusel2017GANsTB} and MUSIQ \cite{Ke2021MUSIQMI} are employed to assess the perceptual quality of the generated images, while CLIP-Text and CLIP-Image \cite{Radford2021LearningTV} scores measure the semantic consistency with the input text and reference image, respectively. To evaluate controllability, we compute the similarity between the generated image and the conditioning input. Specifically, for tasks conditioned on Canny or HED edge maps, we use the Hausdorff Distance (HDD) \cite{Huttenlocher1993ComparingIU,Cao2025RelaCtrlRE} to quantify alignment. For depth-conditioned tasks, we adopt Mean Squared Error (MSE).

\paragraph{Implementation Details.}
We train our model on a machine equipped with 8 NVIDIA H100 GPUs (80GB each), using a batch size of 1 with gradient accumulation of 4, resulting in an effective batch size of 32. Training is performed with the AdamW optimizer using a fixed learning rate of 0.0001 and a weight decay of 0.01. The model is trained for one epoch on the Text-to-Image-2M dataset \cite{Tan2024OminiControlMA}. During preprocessing, we resize the shorter side of each image to 512 pixels and apply a center crop to obtain images of size 512×512. Both image and text dropout rates are set to 0.1. At inference time, we use a 24-step sampling schedule based on the Flow Matching Euler Discrete method \cite{lipman2022flow}, with a guidance scale of 3.5. The random seed is fixed to 42 to ensure reproducibility.


\subsection{Quantitative Analysis}
We conduct a comprehensive evaluation of our method by comparing it with several state-of-the-art (SOTA) controllable generation approaches, including Flux-ControlNet \cite{flux2024}, InstantX-ControlNet \cite{instantx_flux_controlnet_union_2024}, Shakker-Union \cite{shakker_flux_controlnet_union_pro_2025}, XLab-ControlNet-v3 \cite{xlabs_flux_controlnet_2023} and OminiControl \cite{Tan2024OminiControlMA}. The comparison is carried out across four conditional tasks: Canny, Depth, Colorization and HED. We select 5,000 images from the COCO 2017 validation set, resize and center-crop them to 512×512, and generate the corresponding conditional inputs using task-specific preprocessing functions. For the calculation of metrics at 1024 resolution, we selected 500 images from the COCO2017 validation set, processed them into 1024 resolution, and further processed them into specific conditions.

As shown in Table \ref{tab:compare}, we evaluate the quality of the generated images from three perspectives: controllability, visual quality, and semantic consistency.
In terms of controllability, our method achieves the best performance on Canny, Depth, and HED tasks, and shows only a minor difference compared to OminiControl in the Colorization task.
Regarding image quality, our method achieves either the best or second-best scores across all four tasks under FID and MUSIQ metrics.
For consistency, we perform competitively with other top-performing methods in terms of the CLIP-Text Score across all tasks, while showing a clear advantage in CLIP-Image Score, indicating better alignment with the original image structure.In the last row of Table 1, we present the comparison results of the canny-conditioned generation capabilities of different models at a resolution of 1024. We still achieve the best performance in most of the indicators. 

It is worth noting that the model performance needs to be comprehensively evaluated from both the control indicators and the generation quality indicators, as there are two scenarios to consider: First, if the model hardly learns any information, it can generate images with high quality (after all, the backbone is Flux), but its control ability is too weak to meet user requirements. Second, if the model directly incorporates the conditional information into the generated images, the control indicators can be very high (almost directly pasting the conditional map into the generated image and adding some colors), but this will result in very poor image quality, making the model still unusable.

\subsection{Qualitative Analysis}
Figure \ref{fig:multi-subfigures} presents qualitative comparisons between our method and existing approaches. As illustrated, our model consistently produces high-quality, semantically aligned, and visually coherent images across various control tasks. The generated outputs not only preserve the structural cues provided by the conditional inputs, but also exhibit fine-grained details and stylistic fidelity that align well with the input prompts. (Additional qualitative results are available in the Appendix for further comparison.)

These results, in combination with the quantitative metrics, demonstrate that our approach achieves a strong balance between control fidelity, visual realism, and semantic consistency. This highlights the effectiveness and generalizability of our method in handling diverse control modalities while maintaining superior generation quality.

\subsection{Model Efficiency Analysis}
We further analyze the model efficiency of NanoControl in comparison to existing state-of-the-art methods. As shown in Table \ref{tab:compare}, our model significantly reduces the number of parameters compared to traditional ControlNet-based architectures, increasing the parameter count by only 0.024\% relative to the Flux backbone. Even compared with the lightweight OminiControl, our model contains only one-four of its parameters, clearly demonstrating its compactness advantage. Similarly, in terms of computational cost, our method requires substantially fewer FLOPs, increasing only 0.029\% FLOPs relative to the Flux backbone, making it far more efficient than both conventional ControlNet variants and OminiControl.

\begin{table}[t]
    \centering 
    \resizebox{\columnwidth}{!}{%
    \begin{tabular}{lccc} 
        \toprule
        Models & Parameters & FLOPs (512) & FLOPs (1024) \\
        \midrule
        \textit{Flux-base}     & \textit{12 B}            & \textit{9,926 G}    & \textit{29,758 G}  \\
        InstantX      & +1,792 M        & +926 G     & +2,811 G         \\
        Shakker       & +3,302 M        & +2,764 G   & +8,419 G        \\
        Xlab          & +744 M          & +375 G     & +1,134 G        \\
        OminiControl         & +14 M           & +6,639 G   & +26,539 G        \\
        \textbf{Ours} & \textbf{+3 M}            & \textbf{+3 G}      & \textbf{+11 G}        \\ 
        \bottomrule
    \end{tabular}
    }
    \caption{Comparison of model parameters and flops increment. FLOPs(512) refers to the results on 512×512-sized images, while FLOPs(1024) refers to the results on 1024×1024-sized images.}
    \label{tab:model_comparison} 
\end{table}

In summary, our method achieves a strong balance between generation quality and computational efficiency. While delivering superior visual results and quantitative metrics, it effectively reduces parameter size and inference overhead, making it better suited for real-world applications and deployment in resource-constrained environments.



\subsection{Ablation Study}
\paragraph{LoRA rank} We conduct ablation studies on the LoRA rank settings in two key components of our model: the LoRA-style Control Module and the image projection MLP module. Table \ref{tab:ablation}(a) presents the results for the LoRA-style Control Module under different LoRA ranks. As the rank increases from 1 to 32, we observe a general improvement in performance across all metrics, though the gains become progressively marginal. Considering the trade-off between computational cost and generation quality, we choose a rank of 4 as a practical and efficient configuration.
Table \ref{tab:ablation}(b) reports the results for the image projection MLP module. In this case, the model maintains relatively stable performance when the LoRA rank is 32 or higher. However, when the rank drops below 16, there is a noticeable decline in performance metrics. Therefore, we set the rank to 32 in this component to ensure stable and reliable results.

\begin{table*}[t]
  \centering
  \scriptsize
  \hspace{-0.025\textwidth}
  \begin{subtable}[t]{0.3\textwidth}
    \centering
    \begin{tabular}{cccc}
      \toprule
      rank & Hdd $\downarrow$ & FID $\downarrow$ & CLIP-I $\uparrow$\\
      \midrule
      2  & 113.55 & 73.73 & 0.797\\
      \textbf{4}  & \textbf{111.26} & \textbf{70.95} & \textbf{0.809}\\
      8  & 110.84 & 70.01 & 0.810\\
      16 & 109.57 & 69.78 & 0.815\\
      32 & 108.05 & 70.99 & 0.811\\
      \bottomrule
    \end{tabular}
    \vspace{0.5em}
    \caption*{(a)}
  \end{subtable}
  \hspace{0.025\textwidth}
  \begin{subtable}[t]{0.3\textwidth}
    \centering
    \begin{tabular}{cccc}
      \toprule
      rank & Hdd $\downarrow$ & FID $\downarrow$ & CLIP-I $\uparrow$\\
      \midrule
      4  & 113.30 & 72.37 & 0.793\\
      8  & 110.83 & 71.85 & 0.803\\
      16 & 110.48 & 71.58 & 0.807\\
      \textbf{32} & \textbf{111.26} & \textbf{70.95} & \textbf{0.809}\\
      64 & 111.32 & 71.63 & 0.805\\
      \bottomrule
    \end{tabular}
    \vspace{0.5em}
    \caption*{(b)}
  \end{subtable}
  \hspace{0.025\textwidth}
  \begin{subtable}[t]{0.3\textwidth}
    \centering
    \begin{tabular}{cccc}
      \toprule
      method & Hdd $\downarrow$ & FID $\downarrow$ & CLIP-I $\uparrow$\\
      \midrule
      \textbf{KV-Context}  & \textbf{111.26} & \textbf{70.95} & \textbf{0.809}\\
      add  & 113.49 & 72.99 & 0.800\\
      \bottomrule
    \end{tabular}
    \vspace{0.5em}
    \caption*{(c)}
  \end{subtable}
\caption{Ablation Studies. (a) Rank setting in LoRA-style Control Module. (b) Rank setting in image projection MLP module. (c) Different conditioning strategies.}
\label{tab:ablation}
\end{table*}

\paragraph{Conditioning Strategy} 

Most existing ControlNet approaches based on the DiT architecture incorporate condition information by adding it directly to the \texttt{hidden\_states}. To better understand the impact of this design choice, we also evaluate a variant of our method that adopts this additive strategy. Specifically, instead of concatenating \texttt{cond\_key} and \texttt{cond\_value} with the original \texttt{key} and \texttt{value}, we apply cross-attention only with the \texttt{query}, and use the condition features via simple addition.
As shown in Table \ref{tab:ablation}(c), this alternative approach yields inferior performance compared to our proposed method, demonstrating the effectiveness of our KV-Context Augmentation Mechanism design and the importance of structured condition integration.

\section{Conclusion}
\label{conclusion}
NanoControl employs an extremely minimal independent branch as a plug-in integrated into the DIT base model. With only an additional 0.024\% parameter volume and 0.029\% Gflops, it achieves state-of-the-art (SOTA) control capabilities and conditional generation effects. We have demonstrated the effectiveness of NanoControl through extensive quantitative and qualitative experiments. However, we have only validated the approach on spatially aligned tasks and have not yet conducted experiments on spatially misaligned tasks, which will be addressed in our subsequent research. In the future, we will expand NanoControl to text-to-video models, providing more powerful tools for the open-source community.

\bibliography{aaai2026}

\begin{thebibliography}{35}
\providecommand{\natexlab}[1]{#1}

\bibitem[{Cao et~al.(2025)Cao, Wang, Ma, Feng, Zhang, He, Liu, Cheng, Leng, Yin, and Zhang}]{Cao2025RelaCtrlRE}
Cao, K.; Wang, J.; Ma, A.; Feng, J.; Zhang, Z.; He, X.; Liu, S.; Cheng, B.; Leng, D.; Yin, Y.; and Zhang, J. 2025.
\newblock RelaCtrl: Relevance-Guided Efficient Control for Diffusion Transformers.
\newblock \emph{ArXiv}, abs/2502.14377.

\bibitem[{Chen et~al.(2024)Chen, Wu, Luo, Xie, Paul, Luo, Zhao, and Li}]{chen2024pixart}
Chen, J.; Wu, Y.; Luo, S.; Xie, E.; Paul, S.; Luo, P.; Zhao, H.; and Li, Z. 2024.
\newblock Pixart-$\{$$\backslash$delta$\}$: Fast and controllable image generation with latent consistency models.
\newblock \emph{arXiv preprint arXiv:2401.05252}.

\bibitem[{Chen et~al.(2023)Chen, Yu, Ge, Yao, Xie, Wu, Wang, Kwok, Luo, Lu, and Li}]{Chen2023PixArtFT}
Chen, J.; Yu, J.; Ge, C.; Yao, L.; Xie, E.; Wu, Y.; Wang, Z.; Kwok, J.~T.; Luo, P.; Lu, H.; and Li, Z. 2023.
\newblock PixArt-$\alpha$: Fast Training of Diffusion Transformer for Photorealistic Text-to-Image Synthesis.
\newblock \emph{ArXiv}, abs/2310.00426.

\bibitem[{Heusel et~al.(2017)Heusel, Ramsauer, Unterthiner, Nessler, and Hochreiter}]{Heusel2017GANsTB}
Heusel, M.; Ramsauer, H.; Unterthiner, T.; Nessler, B.; and Hochreiter, S. 2017.
\newblock GANs Trained by a Two Time-Scale Update Rule Converge to a Local Nash Equilibrium.
\newblock In \emph{Neural Information Processing Systems}.

\bibitem[{Ho, Jain, and Abbeel(2020)}]{ho2020denoising}
Ho, J.; Jain, A.; and Abbeel, P. 2020.
\newblock Denoising diffusion probabilistic models.
\newblock \emph{Advances in neural information processing systems}, 33: 6840--6851.

\bibitem[{Hu et~al.(2021)Hu, Shen, Wallis, Allen-Zhu, Li, Wang, and Chen}]{Hu2021LoRALA}
Hu, J.~E.; Shen, Y.; Wallis, P.; Allen-Zhu, Z.; Li, Y.; Wang, S.; and Chen, W. 2021.
\newblock LoRA: Low-Rank Adaptation of Large Language Models.
\newblock \emph{ArXiv}, abs/2106.09685.

\bibitem[{Huttenlocher, Klanderman, and Rucklidge(1993)}]{Huttenlocher1993ComparingIU}
Huttenlocher, D.~P.; Klanderman, G.~A.; and Rucklidge, W. 1993.
\newblock Comparing Images Using the Hausdorff Distance.
\newblock \emph{IEEE Trans. Pattern Anal. Mach. Intell.}, 15: 850--863.

\bibitem[{InstantX(2024)}]{instantx_flux_controlnet_union_2024}
InstantX. 2024.
\newblock FLUX.1-dev-Controlnet-Union.

\bibitem[{Ke et~al.(2021)Ke, Wang, Wang, Milanfar, and Yang}]{Ke2021MUSIQMI}
Ke, J.; Wang, Q.; Wang, Y.; Milanfar, P.; and Yang, F. 2021.
\newblock MUSIQ: Multi-scale Image Quality Transformer.
\newblock \emph{2021 IEEE/CVF International Conference on Computer Vision (ICCV)}, 5128--5137.

\bibitem[{Labs(2024)}]{flux2024}
Labs, B.~F. 2024.
\newblock FLUX.
\newblock \url{https://github.com/black-forest-labs/flux}.

\bibitem[{Li, Li, and Hoi(2023)}]{Li2023BLIPDiffusionPS}
Li, D.; Li, J.; and Hoi, S. C.~H. 2023.
\newblock BLIP-Diffusion: Pre-trained Subject Representation for Controllable Text-to-Image Generation and Editing.
\newblock \emph{ArXiv}, abs/2305.14720.

\bibitem[{Li et~al.(2024)Li, Cheng, Chen, Sun, Shen, Ran, Chen, Liu, and Wang}]{Li2024ControlARCI}
Li, Z.; Cheng, T.; Chen, S.; Sun, P.; Shen, H.; Ran, L.; Chen, X.; Liu, W.; and Wang, X. 2024.
\newblock ControlAR: Controllable Image Generation with Autoregressive Models.
\newblock \emph{ArXiv}, abs/2410.02705.

\bibitem[{Lipman et~al.(2022)Lipman, Chen, Ben-Hamu, Nickel, and Le}]{lipman2022flow}
Lipman, Y.; Chen, R.~T.; Ben-Hamu, H.; Nickel, M.; and Le, M. 2022.
\newblock Flow matching for generative modeling.
\newblock \emph{arXiv preprint arXiv:2210.02747}.

\bibitem[{Lopez et~al.(2020)Lopez, Boyeau, Yosef, Jordan, and Regier}]{Lopez2020AUTOENCODINGVB}
Lopez, R.; Boyeau, P.; Yosef, N.; Jordan, M.~I.; and Regier, J. 2020.
\newblock AUTO-ENCODING VARIATIONAL BAYES.

\bibitem[{Mao et~al.(2025)Mao, Zhang, Pan, Jiang, Han, Liu, and Zhou}]{mao2025ace++}
Mao, C.; Zhang, J.; Pan, Y.; Jiang, Z.; Han, Z.; Liu, Y.; and Zhou, J. 2025.
\newblock Ace++: Instruction-based image creation and editing via context-aware content filling.
\newblock \emph{arXiv preprint arXiv:2501.02487}.

\bibitem[{Mou et~al.(2023)Mou, Wang, Xie, Zhang, Qi, Shan, and Qie}]{Mou2023T2IAdapterLA}
Mou, C.; Wang, X.; Xie, L.; Zhang, J.; Qi, Z.; Shan, Y.; and Qie, X. 2023.
\newblock T2I-Adapter: Learning Adapters to Dig out More Controllable Ability for Text-to-Image Diffusion Models.
\newblock In \emph{AAAI Conference on Artificial Intelligence}.

\bibitem[{Peebles and Xie(2023)}]{peebles2023scalable}
Peebles, W.; and Xie, S. 2023.
\newblock Scalable diffusion models with transformers.
\newblock In \emph{Proceedings of the IEEE/CVF international conference on computer vision}, 4195--4205.

\bibitem[{Peebles and Xie(2022)}]{Peebles2022ScalableDM}
Peebles, W.~S.; and Xie, S. 2022.
\newblock Scalable Diffusion Models with Transformers.
\newblock \emph{2023 IEEE/CVF International Conference on Computer Vision (ICCV)}, 4172--4182.

\bibitem[{Peng et~al.(2024)Peng, Wang, Zhang, Li, Yang, and Jia}]{Peng2024ControlNeXtPA}
Peng, B.; Wang, J.; Zhang, Y.; Li, W.; Yang, M.; and Jia, J. 2024.
\newblock ControlNeXt: Powerful and Efficient Control for Image and Video Generation.
\newblock \emph{ArXiv}, abs/2408.06070.

\bibitem[{Podell et~al.(2023)Podell, English, Lacey, Blattmann, Dockhorn, M{\"u}ller, Penna, and Rombach}]{podell2023sdxl}
Podell, D.; English, Z.; Lacey, K.; Blattmann, A.; Dockhorn, T.; M{\"u}ller, J.; Penna, J.; and Rombach, R. 2023.
\newblock Sdxl: Improving latent diffusion models for high-resolution image synthesis.
\newblock \emph{arXiv preprint arXiv:2307.01952}.

\bibitem[{Radford et~al.(2021)Radford, Kim, Hallacy, Ramesh, Goh, Agarwal, Sastry, Askell, Mishkin, Clark, Krueger, and Sutskever}]{Radford2021LearningTV}
Radford, A.; Kim, J.~W.; Hallacy, C.; Ramesh, A.; Goh, G.; Agarwal, S.; Sastry, G.; Askell, A.; Mishkin, P.; Clark, J.; Krueger, G.; and Sutskever, I. 2021.
\newblock Learning Transferable Visual Models From Natural Language Supervision.
\newblock In \emph{International Conference on Machine Learning}.

\bibitem[{Raffel et~al.(2020)Raffel, Shazeer, Roberts, Lee, Narang, Matena, Zhou, Li, and Liu}]{raffel2020exploring}
Raffel, C.; Shazeer, N.; Roberts, A.; Lee, K.; Narang, S.; Matena, M.; Zhou, Y.; Li, W.; and Liu, P.~J. 2020.
\newblock Exploring the limits of transfer learning with a unified text-to-text transformer.
\newblock \emph{Journal of machine learning research}, 21(140): 1--67.

\bibitem[{Ramesh et~al.(2022)Ramesh, Dhariwal, Nichol, Chu, and Chen}]{Ramesh2022HierarchicalTI}
Ramesh, A.; Dhariwal, P.; Nichol, A.; Chu, C.; and Chen, M. 2022.
\newblock Hierarchical Text-Conditional Image Generation with CLIP Latents.
\newblock \emph{ArXiv}, abs/2204.06125.

\bibitem[{Rombach et~al.(2021)Rombach, Blattmann, Lorenz, Esser, and Ommer}]{Rombach2021HighResolutionIS}
Rombach, R.; Blattmann, A.; Lorenz, D.; Esser, P.; and Ommer, B. 2021.
\newblock High-Resolution Image Synthesis with Latent Diffusion Models.
\newblock \emph{2022 IEEE/CVF Conference on Computer Vision and Pattern Recognition (CVPR)}, 10674--10685.

\bibitem[{Rombach et~al.(2022)Rombach, Blattmann, Lorenz, Esser, and Ommer}]{rombach2022high}
Rombach, R.; Blattmann, A.; Lorenz, D.; Esser, P.; and Ommer, B. 2022.
\newblock High-resolution image synthesis with latent diffusion models.
\newblock In \emph{Proceedings of the IEEE/CVF conference on computer vision and pattern recognition}, 10684--10695.

\bibitem[{Ronneberger, Fischer, and Brox(2015)}]{ronneberger2015u}
Ronneberger, O.; Fischer, P.; and Brox, T. 2015.
\newblock U-net: Convolutional networks for biomedical image segmentation.
\newblock In \emph{Medical image computing and computer-assisted intervention--MICCAI 2015: 18th international conference, Munich, Germany, October 5-9, 2015, proceedings, part III 18}, 234--241. Springer.

\bibitem[{Saharia et~al.(2022)Saharia, Chan, Saxena, Li, Whang, Denton, Ghasemipour, Gontijo~Lopes, Karagol~Ayan, Salimans et~al.}]{saharia2022photorealistic}
Saharia, C.; Chan, W.; Saxena, S.; Li, L.; Whang, J.; Denton, E.~L.; Ghasemipour, K.; Gontijo~Lopes, R.; Karagol~Ayan, B.; Salimans, T.; et~al. 2022.
\newblock Photorealistic text-to-image diffusion models with deep language understanding.
\newblock \emph{Advances in neural information processing systems}, 35: 36479--36494.

\bibitem[{Shakker-Labs(2025)}]{shakker_flux_controlnet_union_pro_2025}
Shakker-Labs. 2025.
\newblock FLUX.1-dev-ControlNet-Union-Pro.

\bibitem[{Song et~al.(2020)Song, Sohl-Dickstein, Kingma, Kumar, Ermon, and Poole}]{song2020score}
Song, Y.; Sohl-Dickstein, J.; Kingma, D.~P.; Kumar, A.; Ermon, S.; and Poole, B. 2020.
\newblock Score-based generative modeling through stochastic differential equations.
\newblock \emph{arXiv preprint arXiv:2011.13456}.

\bibitem[{Tan et~al.(2024)Tan, Liu, Yang, Xue, and Wang}]{Tan2024OminiControlMA}
Tan, Z.; Liu, S.; Yang, X.; Xue, Q.; and Wang, X. 2024.
\newblock OminiControl: Minimal and Universal Control for Diffusion Transformer.
\newblock \emph{ArXiv}, abs/2411.15098.

\bibitem[{Vaswani et~al.(2017)Vaswani, Shazeer, Parmar, Uszkoreit, Jones, Gomez, Kaiser, and Polosukhin}]{vaswani2017attention}
Vaswani, A.; Shazeer, N.; Parmar, N.; Uszkoreit, J.; Jones, L.; Gomez, A.~N.; Kaiser, {\L}.; and Polosukhin, I. 2017.
\newblock Attention is all you need.
\newblock \emph{Advances in neural information processing systems}, 30.

\bibitem[{XLabs-AI(2023)}]{xlabs_flux_controlnet_2023}
XLabs-AI. 2023.
\newblock flux-controlnet-collections.

\bibitem[{Ye et~al.(2023)Ye, Zhang, Liu, Han, and Yang}]{Ye2023IPAdapterTC}
Ye, H.; Zhang, J.; Liu, S.; Han, X.; and Yang, W. 2023.
\newblock IP-Adapter: Text Compatible Image Prompt Adapter for Text-to-Image Diffusion Models.
\newblock \emph{ArXiv}, abs/2308.06721.

\bibitem[{Zhang, Rao, and Agrawala(2023)}]{Zhang2023AddingCC}
Zhang, L.; Rao, A.; and Agrawala, M. 2023.
\newblock Adding Conditional Control to Text-to-Image Diffusion Models.
\newblock \emph{2023 IEEE/CVF International Conference on Computer Vision (ICCV)}, 3813--3824.

\bibitem[{Zhao et~al.(2023)Zhao, Chen, Chen, Bao, Hao, Yuan, and Wong}]{Zhao2023UniControlNetAC}
Zhao, S.; Chen, D.; Chen, Y.-C.; Bao, J.; Hao, S.; Yuan, L.; and Wong, K.-Y.~K. 2023.
\newblock Uni-ControlNet: All-in-One Control to Text-to-Image Diffusion Models.
\newblock \emph{ArXiv}, abs/2305.16322.

\end{thebibliography}



\newpage
\appendix
\twocolumn[
\begin{center}
\Large \textbf{Supplementary Material}
\vspace{1cm}
\end{center}
]

\begin{table}[htbp]
    \centering 
    \begin{tabular}{lccc} 
        \toprule
        Models & Hdd $\downarrow$ & FID $\downarrow$     \\
        \midrule
        Flux-base     & 110.28        & 108.04        \\
        InstantX      & 116.96        & 116.77        \\
        Shakker       & 119.15        & 126.65        \\
        Xlab          & 109.97        & 101.19         \\
        OminiControl  & 114.27        & 99.16       \\
        \textbf{Ours}          & \textbf{106.43}        & \textbf{95.70}       \\ 
        \bottomrule
    \end{tabular}
    \caption{Quantitative results of our method and other approaches combined with LoRA.}
    \label{tab:model_comparison_lora} 
\end{table}

\begin{table}[t]
    \centering 
    \begin{tabular}{lccc} 
        \toprule
        Methods & Hdd $\downarrow$ & FID $\downarrow$ & CLIP Image$\uparrow$     \\
        \midrule
        Layer-by-Layer   & 117.85     & 84.19   & 0.752      \\
        \textbf{Ours}       & \textbf{111.26}     & \textbf{70.95}   & \textbf{0.809}      \\ 
        \bottomrule
    \end{tabular}
    \caption{Ablation Study on Conditional Injection.}
    \label{tab:model_comparison_block} 
\end{table}

\section{Evaluation with LoRA integration}
As a plug-and-play model, our approach exhibits strong extensibility. To further evaluate its adaptability, we investigate the effectiveness of combining our method with existing LoRA models, and assess the fusion performance using quantitative metrics. Specifically, we incorporate a PixArt-style LoRA model into our framework and all other methods to generate pixel-art-style images. This LoRA model was obtained from an open-source community website and requires adding the prefix "Pixel Art " to the prompt. We randomly selected 500 images from the COCO test set and conducted experiments based on the Canny Conditioning task, using FID and HDD as evaluation metrics. For FID computation, the reference image set consists of outputs generated by the native Flux model with prompts prefixed by "Pixel Art ". It is used to evaluate the ability of various canny condition models combined with LoRA plugins to maintain the LoRA style. A lower FID indicates better style preservation and stronger capability to integrate with LoRA. Table \ref{tab:model_comparison_lora} presents the comparison results, demonstrating that our method achieves superior performance in both controllability and preserving the LoRA style.

\section{Ablation Study on Conditional Injection}

In this section, we further investigate different strategies for injecting conditional information. In our proposed method, the original conditional signal is directly embedded into the attention layers at each block through our LoRA-style Control Module for information fusion. However, the common approach used in existing conditional control models is to input control information at the network entrance, process it, and then pass it to the next block, and so on. We refer to this approach as layer-by-layer transmission. Table \ref{tab:model_comparison_block} presents the comparison between these two approaches. The results show that our method achieves superior performance, indicating that the more direct injection of conditional information is beneficial for enhancing the model’s controllability.

\section{Additional quantitative results at 1024×1024 resolution}
Table \ref{tab:compare_1024} presents the relevant metric results on the Depth, Coloring, and HED tasks using images at a resolution of 1024×1024, further demonstrating the superiority of the proposed method in terms of controllability and related aspects. Due to the large amount of information in 1024×1024 images, the LoRA rank is adjusted according to different tasks. For example, the LoRA rank of 32 is used for the Coloring task.

\begin{table*}[htbp]
    \centering
    \footnotesize
    \begin{adjustbox}{width=\textwidth}
    \begin{tabular}{ccccccccc}
        \toprule
        \multirow{2}{*}{Task} & \multirow{2}{*}{Methods}
        & \multicolumn{1}{c}{Controllability} & \multicolumn{2}{c}{Image Quality} & \multicolumn{2}{c}{Consistency} \\
        \cmidrule(lr){3-3} \cmidrule(lr){4-5} \cmidrule(lr){6-7}
        & & HDD$\downarrow$ / MSE$\downarrow$ & FID$\downarrow$ & MUSIQ$\uparrow$  & CLIP Text$\uparrow$ & CLIP Image$\uparrow$ \\
        \midrule
        \multirow{6}{*}{\makecell{Depth*}} 
        & Flux-Contolnet            & \textbf{445.01}       & \textbf{75.17}        & \underline{63.91}     & \textbf{0.254}        & \underline{0.777} \\
        & InstantX-Controlnet       & 1455.66               & 116.41                & 50.92                 & 0.226                 & 0.702 \\
        & Shakker-Union             & 1267.40               & 96.28                 & 52.35                 & 0.241                 & 0.734 \\
        & Xlab-Controlnet-v3        & 5690.33               & 109.89                & 61.20                 & \underline{0.252}     & 0.689 \\
        & OminiControl              & 1640.95               & 101.92                & 52.29                 & 0.240                 & 0.705 \\
        & Ours                      & \underline{488.60}    & \underline{84.75}     & \textbf{69.00}        & 0.249                 & \textbf{0.779} \\
        \midrule
        \multirow{4}{*}{\makecell{Coloring*}} 
        & InstantX-Controlnet       & 36.36                 & 41.51             & \underline{50.62}     & 0.248                 & 0.908 \\
        & Shakker-Union             & \underline{27.65}     & \textbf{40.55}    & 49.19                 & \underline{0.251}     & \underline{0.916} \\
        & OminiControl              & 67.98                 & 42.11             & 50.11                 & \underline{0.251}     & 0.912 \\
        & Ours                      & \textbf{21.41}        & \underline{40.94} & \textbf{50.83}        & \textbf{0.253}        & \textbf{0.920} \\
        \midrule
        \multirow{2}{*}{\makecell{Hed*}} 
        & Xlab-Controlnet-v3        & 114.32            & 90.47             & \textbf{64.96}        & \textbf{0.258}        & 0.741 \\
        & Ours                      & \textbf{113.37}   & \textbf{69.74}    & 57.31                 & 0.255                 & \textbf{0.838} \\
        \bottomrule
    \end{tabular}
    \end{adjustbox}
    \caption{Results on the Depth, Coloring, and HED tasks using images at a resolution of 1024×1024. Best results are shown in \textbf{bold}, and second-best results are \underline{underlined}.}
    \label{tab:compare_1024}
\end{table*}

\section{Comparisons with more methods}

In addition to the comparison with models based on the Flux backbone, we also conducted a quantitative evaluation against models built upon the SD backbone. In particular, we compare our approach against ControlNet variants built on SD1.5 and SDXL backbones. The results are presented in Table \ref{tab:compare_sd}, where our method continues to demonstrate strong performance according to the reported metrics. The experiments were conducted at a resolution of 512.

\begin{table*}[htbp]
    \centering
    \footnotesize
    \begin{adjustbox}{width=\textwidth}
    \begin{tabular}{ccccccccc}
        \toprule
        \multirow{2}{*}{Task} & \multirow{2}{*}{Methods}
        & \multicolumn{1}{c}{Controllability} & \multicolumn{2}{c}{Image Quality} & \multicolumn{2}{c}{Consistency} \\
        \cmidrule(lr){3-3} \cmidrule(lr){4-5} \cmidrule(lr){6-7}
        & & HDD$\downarrow$ / MSE$\downarrow$ & FID$\downarrow$ & MUSIQ$\uparrow$  & CLIP Text$\uparrow$ & CLIP Image$\uparrow$ \\
        \midrule
        \multirow{5}{*}{\makecell{Canny}} 
        & SD1.5-Controlnet          & \underline{105.59}& \textbf{15.99}    & \underline{71.29} & 0.250             & 0.790 \\
        & SD1.5-T2I-Adapter         & 109.04 & 17.90    & 68.39 & \underline{0.253}             & 0.763 \\
        & SD1.5-unicontrol          & 106.72& 17.10    & 69.19 & 0.251             & 0.780 \\
        & SDXL-Controlnet-XS        & 110.58            & 18.70             & 63.34             & 0.248    & \underline{0.792} \\
        & Ours                      & \textbf{102.22}   & \underline{16.99} & \textbf{71.86}    & \textbf{0.254} & \textbf{0.793} \\
        \midrule
        \multirow{5}{*}{\makecell{Depth}} 
        & SD1.5-Controlnet          & \underline{776.35}    & \textbf{19.16}    & \underline{70.39}     & 0.252     & \underline{0.735} \\
        & SD1.5-T2I-Adapter         & 1516.98 & 21.36    & 69.04 & \underline{0.253}             & 0.730 \\
        & SD1.5-unicontrol          & 1019.84& 23.75    & 65.85 & \textbf{0.259}             & 0.726 \\
        & SDXL-Controlnet-XS        & 1076.37               & 28.82             & 50.59                 & 0.251                 & 0.723 \\
        & Ours                      & \textbf{567.95}       & \underline{19.48} & \textbf{71.92}        & \underline{0.253}        & \textbf{0.759} \\
        \bottomrule
    \end{tabular}
    \end{adjustbox}
    \caption{Quantitative evaluation in comparison with the SD series models. Best results are shown in \textbf{bold}, and second-best results are \underline{underlined}.}
    \label{tab:compare_sd}
\end{table*}

\section{More Visualized Results}
\begin{figure*}[htbp]
    \centering
    \setlength{\tabcolsep}{1pt} 

    \begin{tabular}{cccccccc}
        \scriptsize Canny & 
        \scriptsize  & 
        \scriptsize Depth & 
        \scriptsize  & 
        \scriptsize Color & 
        \scriptsize  & 
        \scriptsize Hed &
        \scriptsize  \\

        \includegraphics[width=0.12\textwidth]{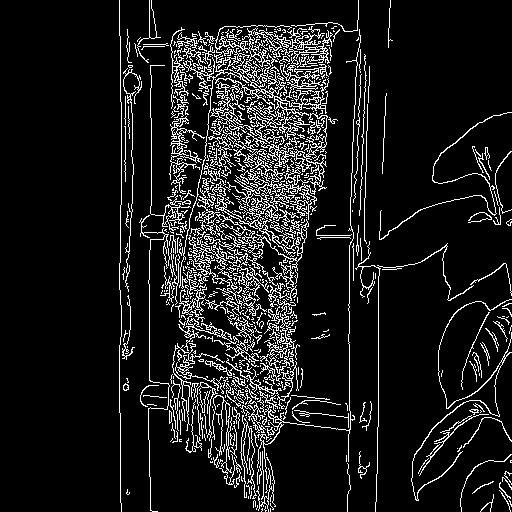} &
        \includegraphics[width=0.12\textwidth]{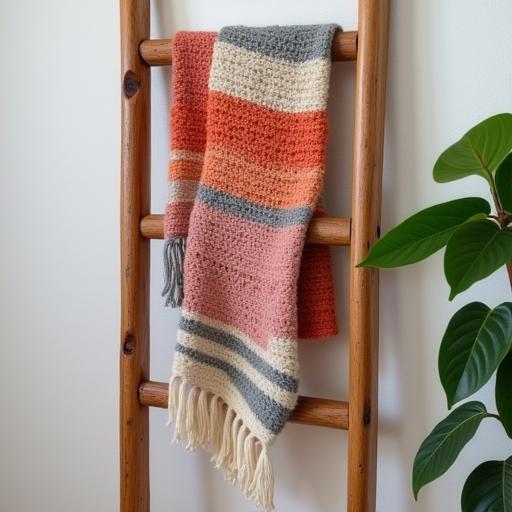} &
        \includegraphics[width=0.12\textwidth]{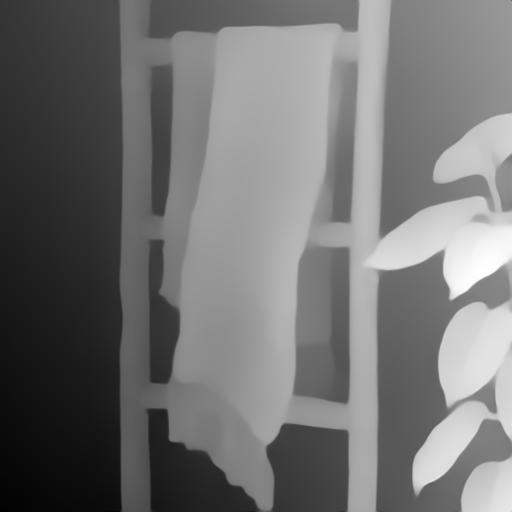} &
        \includegraphics[width=0.12\textwidth]{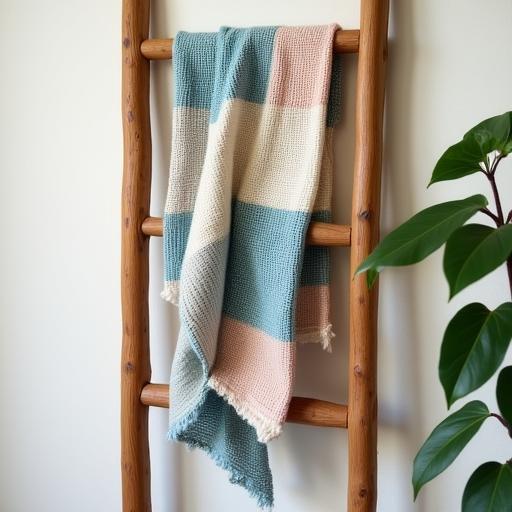} &
        \includegraphics[width=0.12\textwidth]{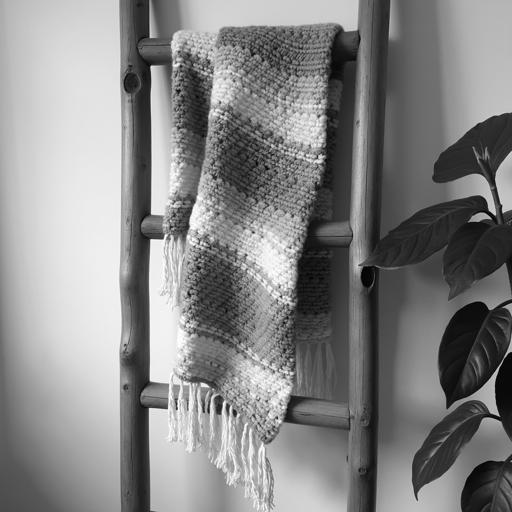} &
        \includegraphics[width=0.12\textwidth]{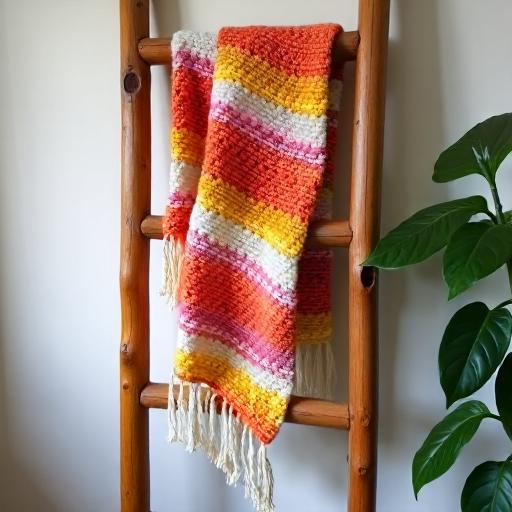} &
        \includegraphics[width=0.12\textwidth]{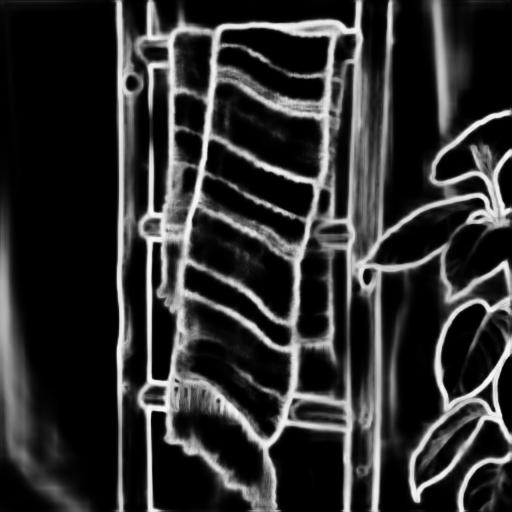} &
        \includegraphics[width=0.12\textwidth]{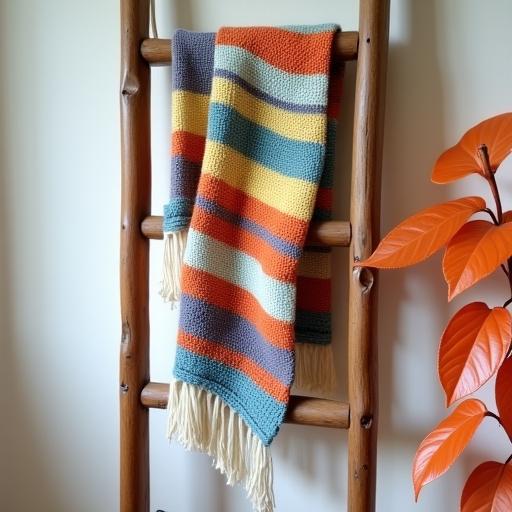} \\
        \multicolumn{8}{l}{\scriptsize \parbox[t]{0.95\textwidth} {A wooden ladder with a hand-woven, multi-colored blanket hanging from it. The blanket is made of yarn and has a unique design, making it an eye-catching piece. The ladder is placed against a wall, and the blanket is draped over it, creating a cozy and inviting atmosphere. The combination of the wooden ladder and the hand-woven blanket adds a rustic touch to the scene.}} \\

        \includegraphics[width=0.12\textwidth]{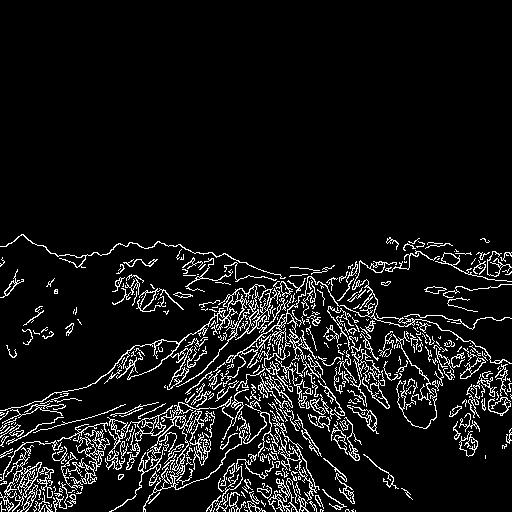} &
        \includegraphics[width=0.12\textwidth]{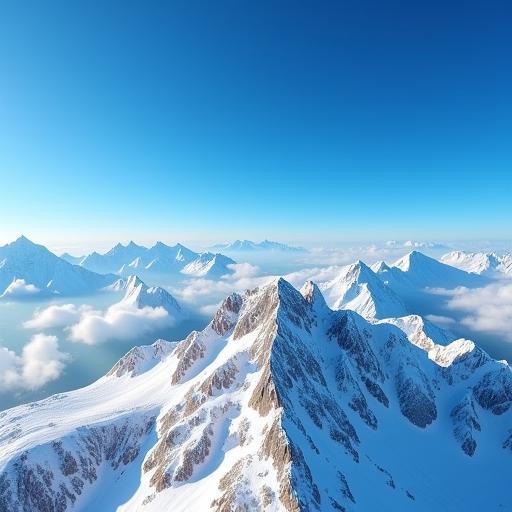} &
        \includegraphics[width=0.12\textwidth]{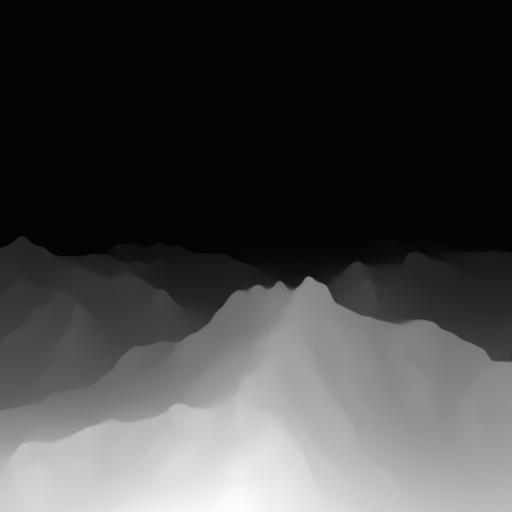} &
        \includegraphics[width=0.12\textwidth]{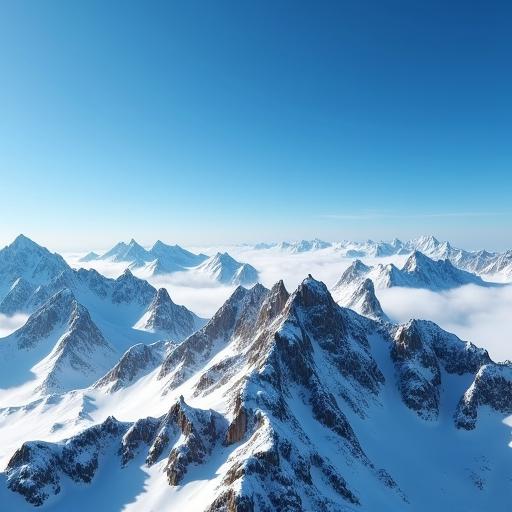} &
        \includegraphics[width=0.12\textwidth]{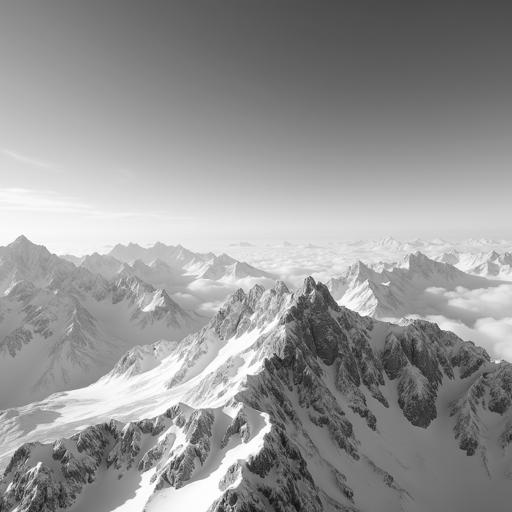} &
        \includegraphics[width=0.12\textwidth]{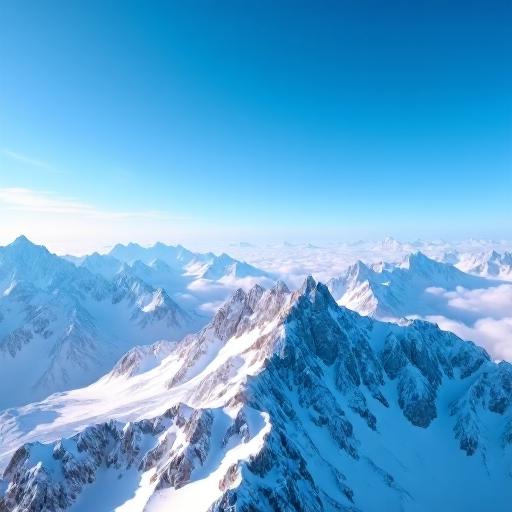} &
        \includegraphics[width=0.12\textwidth]{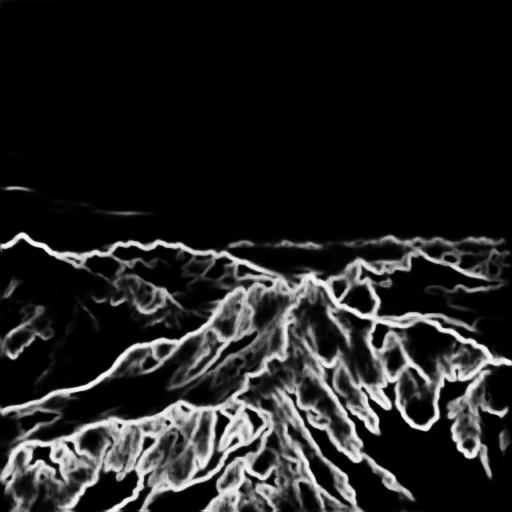} &
        \includegraphics[width=0.12\textwidth]{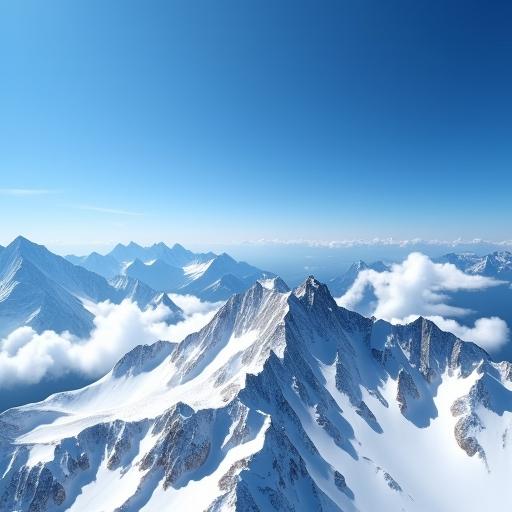} \\
        \multicolumn{8}{l}{\scriptsize \parbox[t]{0.95\textwidth} {A breathtaking view of a mountainous landscape with a clear blue sky above. The mountains are covered in snow, creating a picturesque scene. The sky is filled with clouds, adding depth and beauty to the landscape. The mountains are visible in the distance, with some appearing closer and others further away. The scene is serene and captivating, showcasing the beauty of nature.}} \\

        \includegraphics[width=0.12\textwidth]{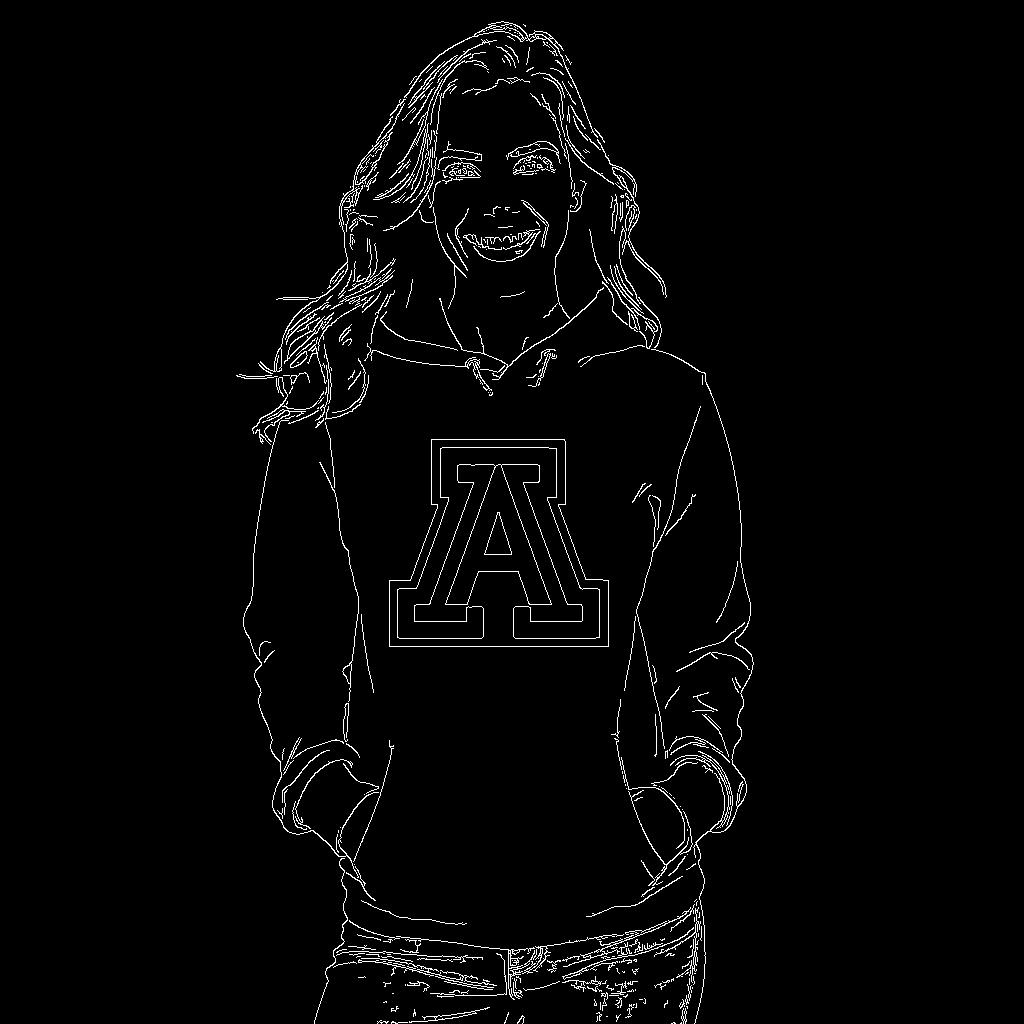} &
        \includegraphics[width=0.12\textwidth]{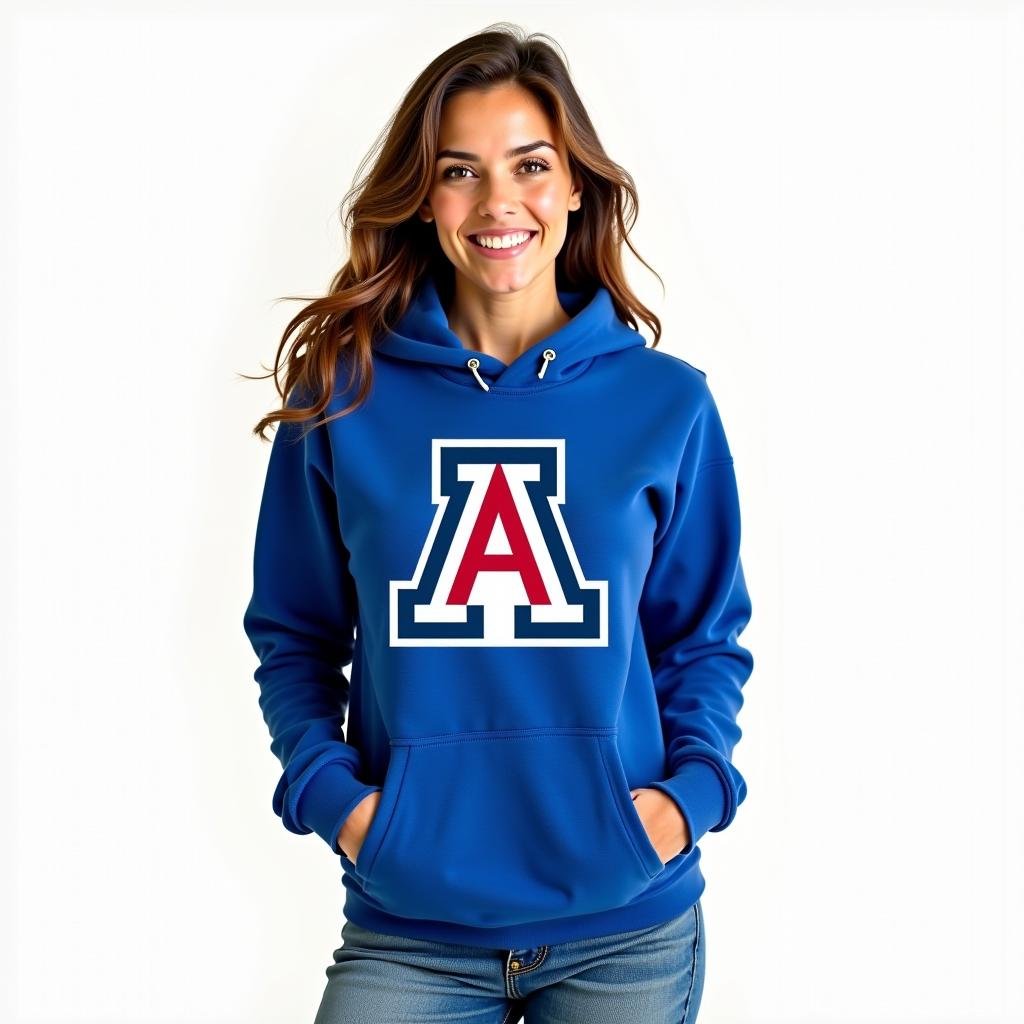} &
        \includegraphics[width=0.12\textwidth]{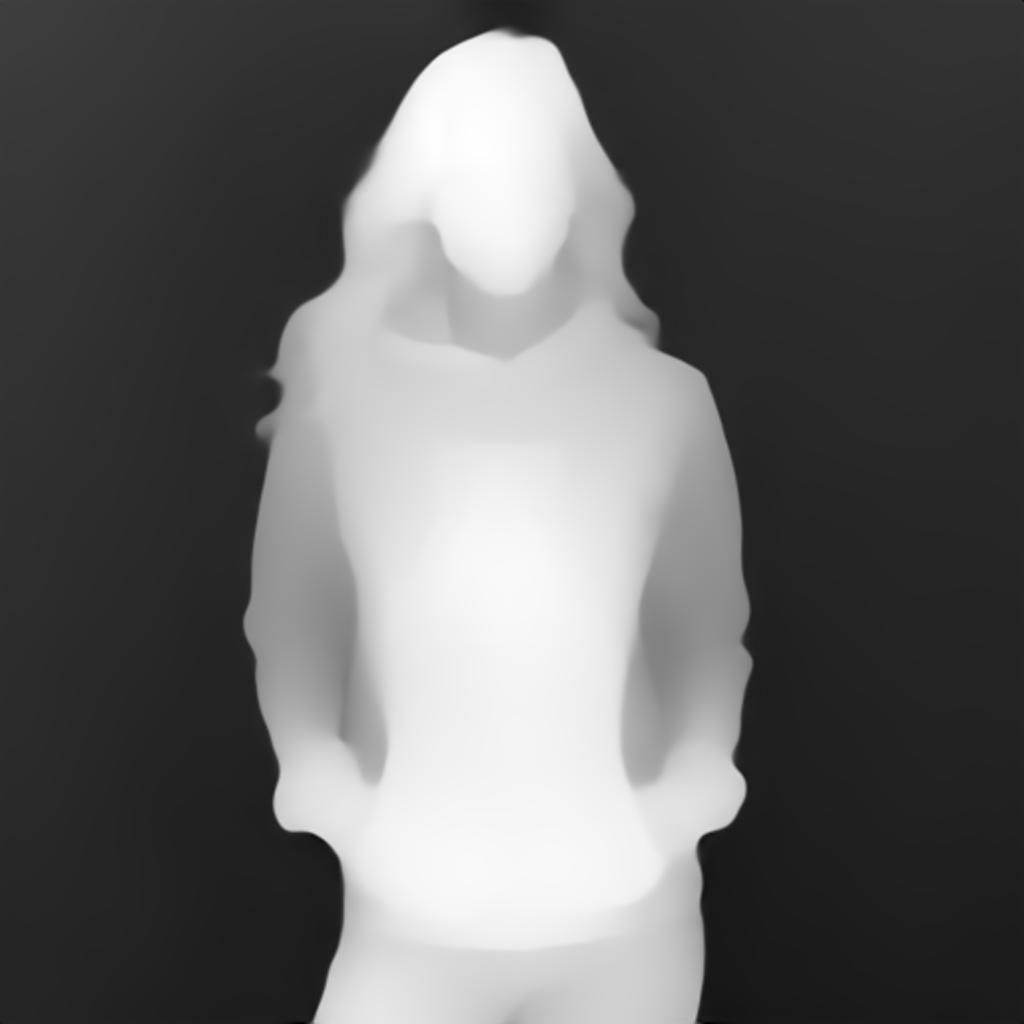} &
        \includegraphics[width=0.12\textwidth]{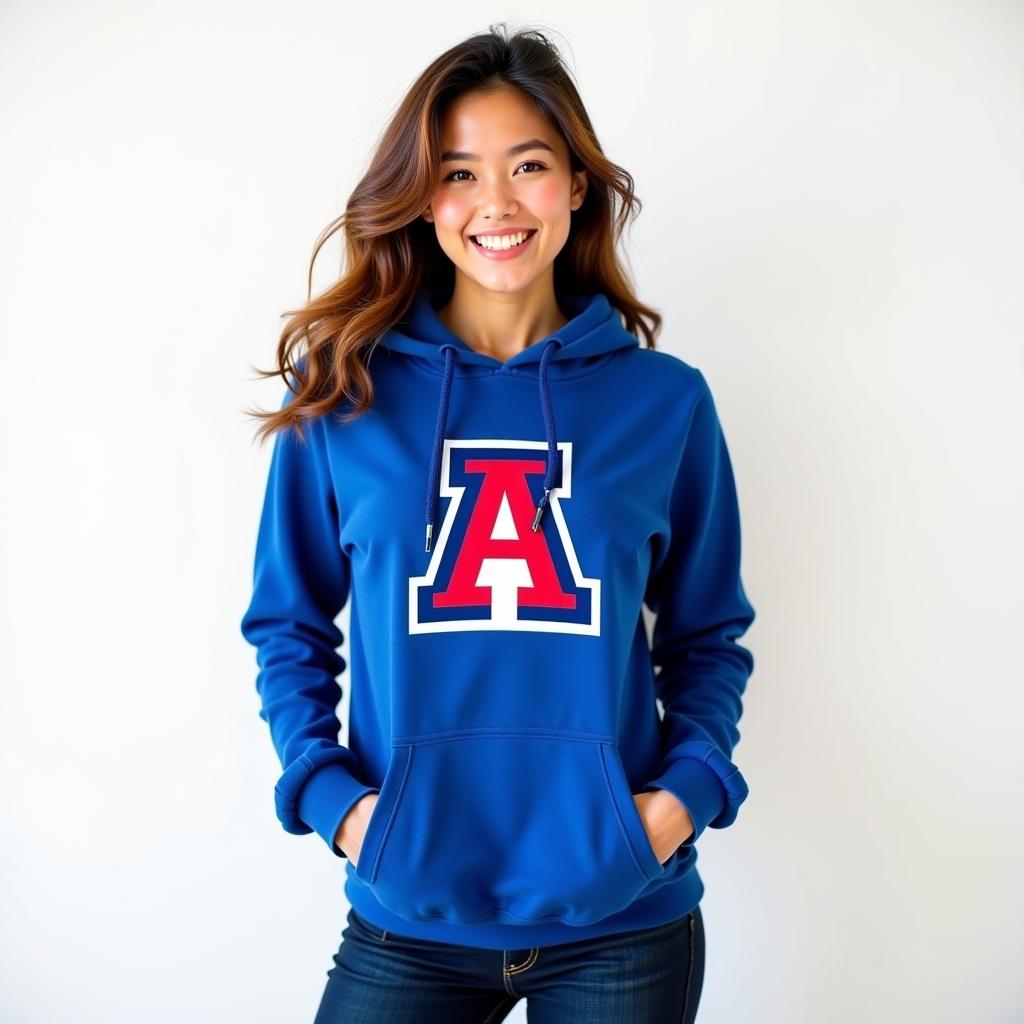} &
        \includegraphics[width=0.12\textwidth]{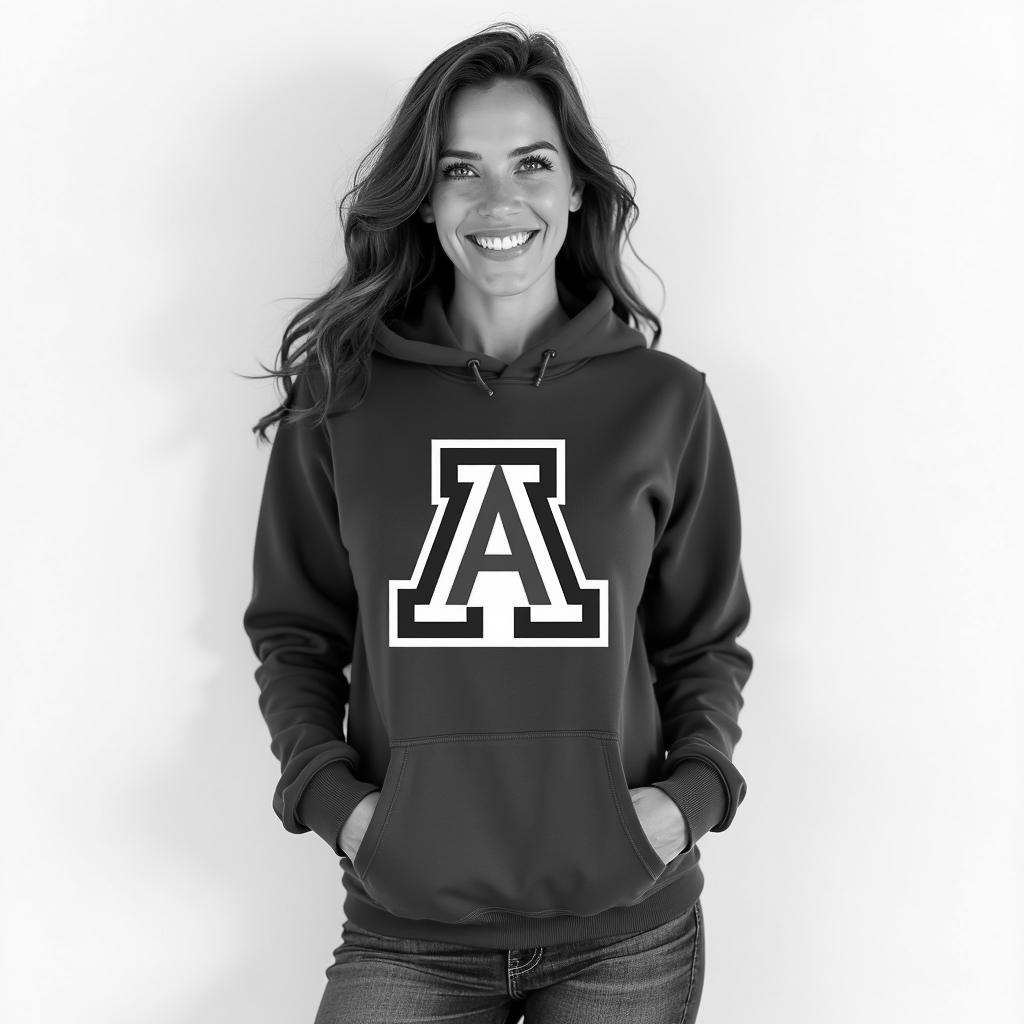} &
        \includegraphics[width=0.12\textwidth]{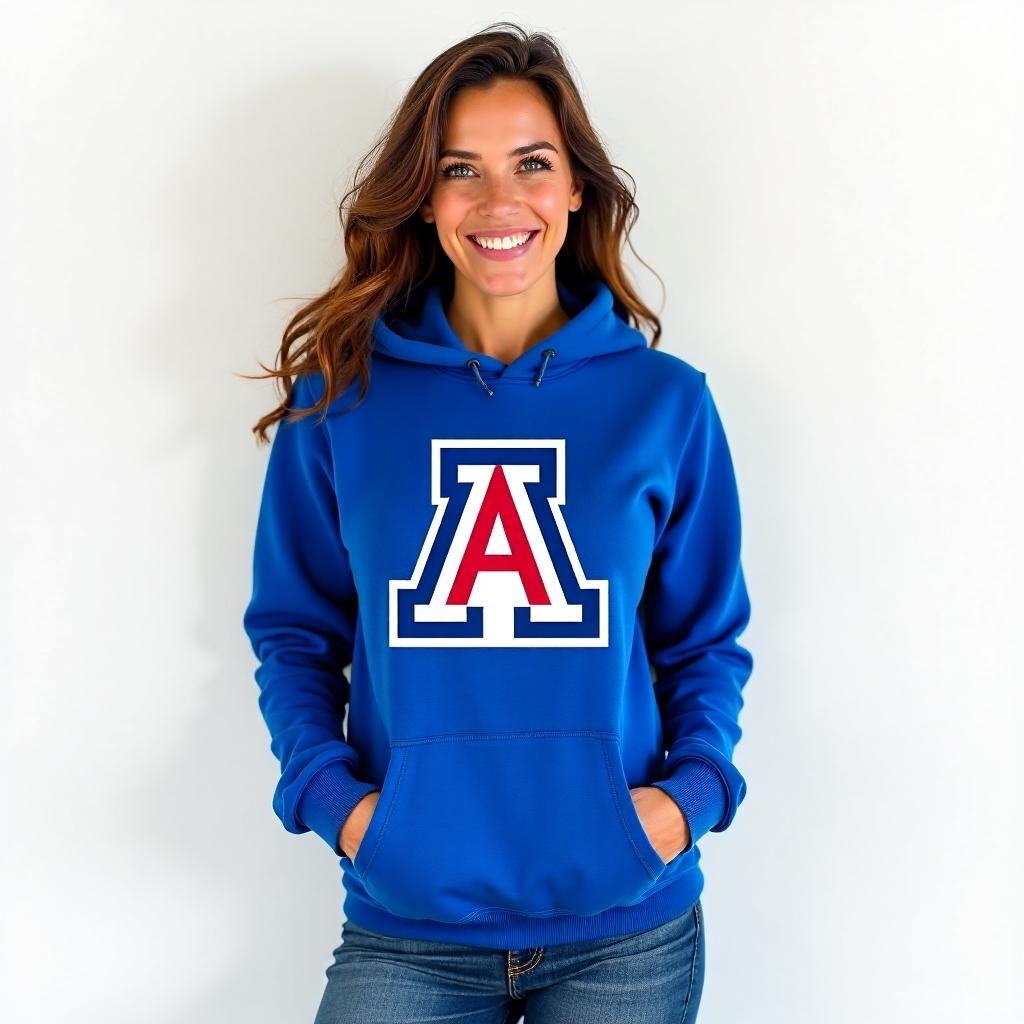} &
        \includegraphics[width=0.12\textwidth]{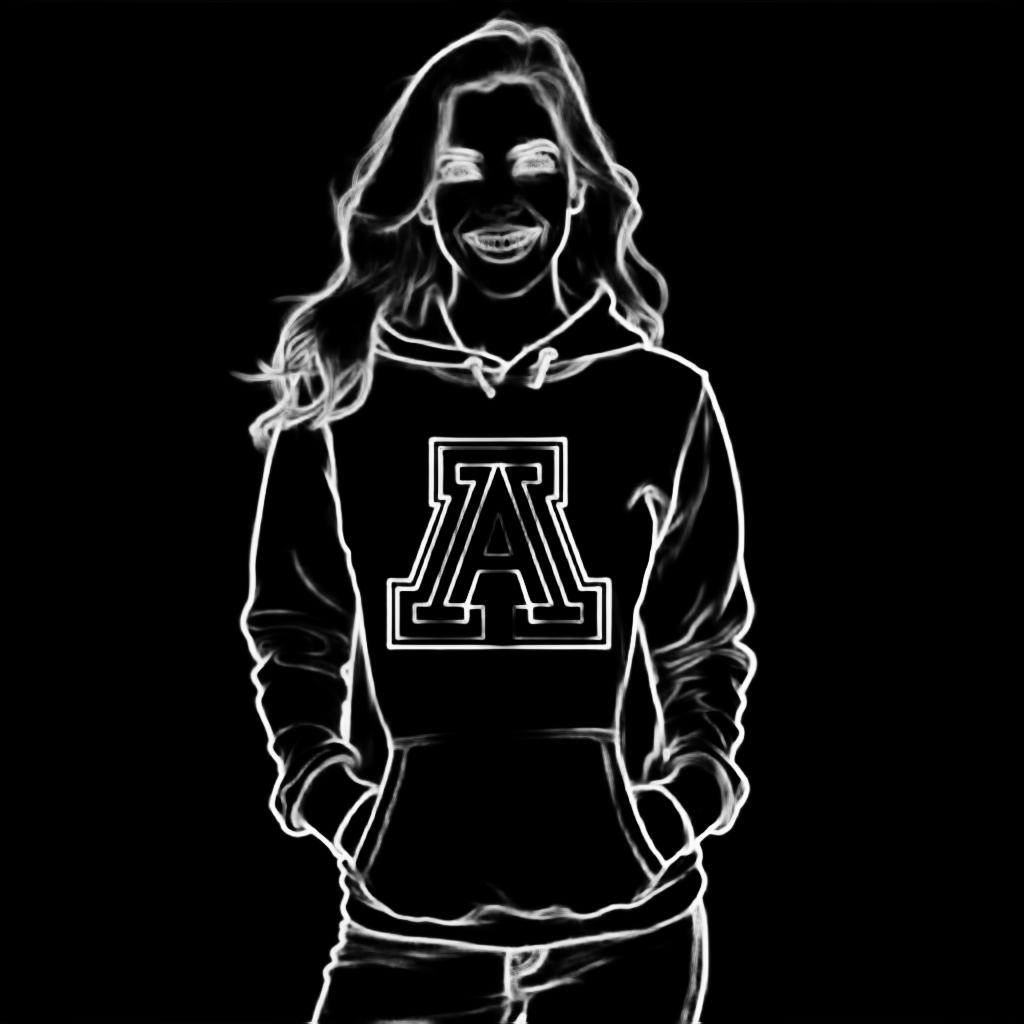} &
        \includegraphics[width=0.12\textwidth]{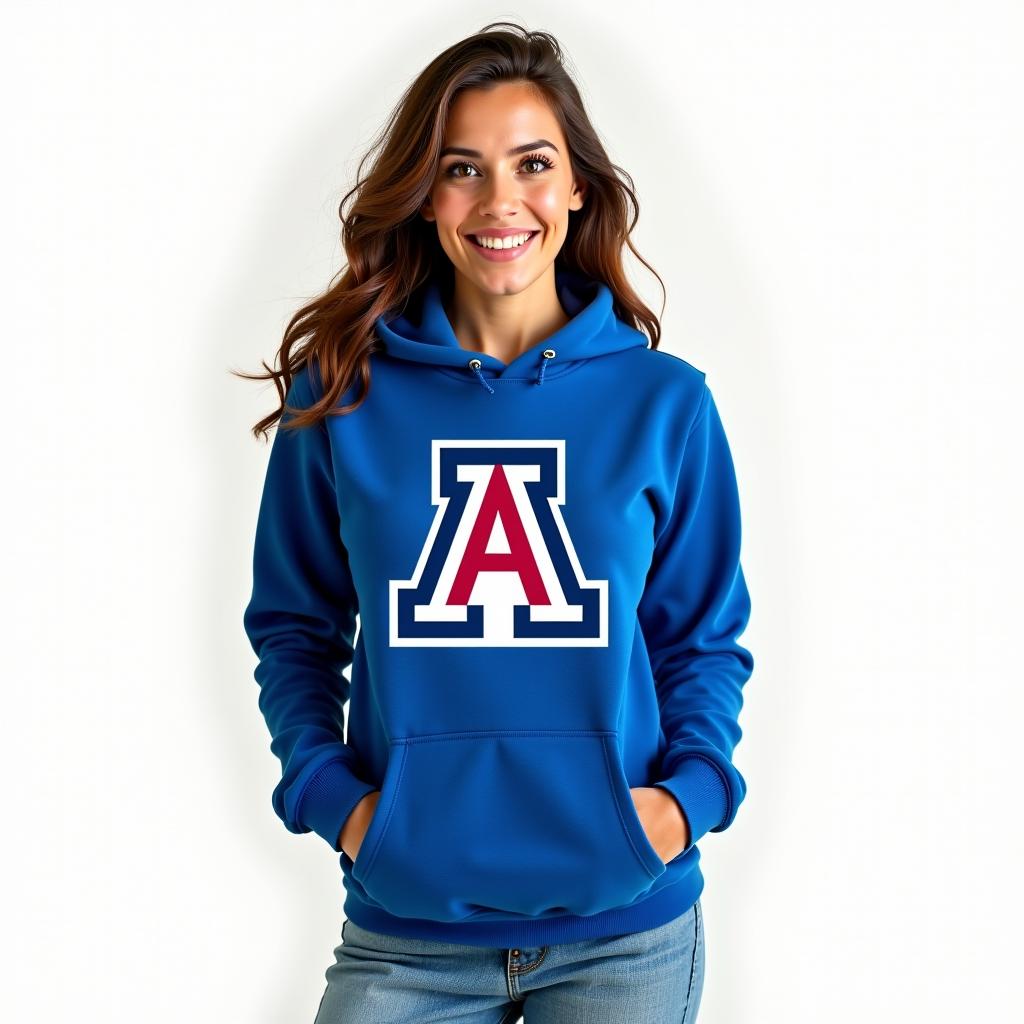} \\
        \multicolumn{8}{l}{\scriptsize \parbox[t]{0.95\textwidth} {A woman wearing a blue hoodie with the letters "A" and "U" on it, representing the University of Arizona. She is standing in front of a white background, posing for the camera. The hoodie is a sweatshirt, and the woman appears to be smiling, showcasing her pride in her university. The hoodie is a popular choice for college students, as it is comfortable and easy to wear.}} \\

        \includegraphics[width=0.12\textwidth]{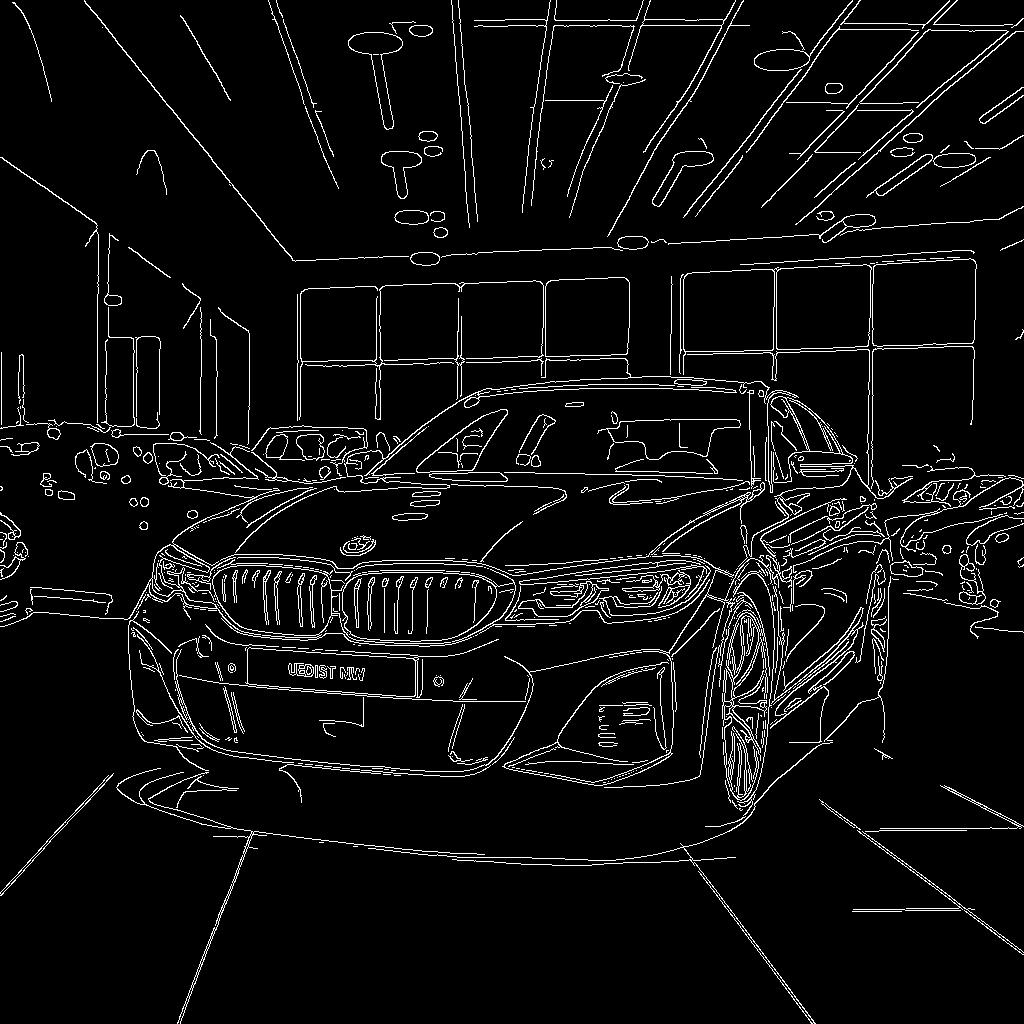} &
        \includegraphics[width=0.12\textwidth]{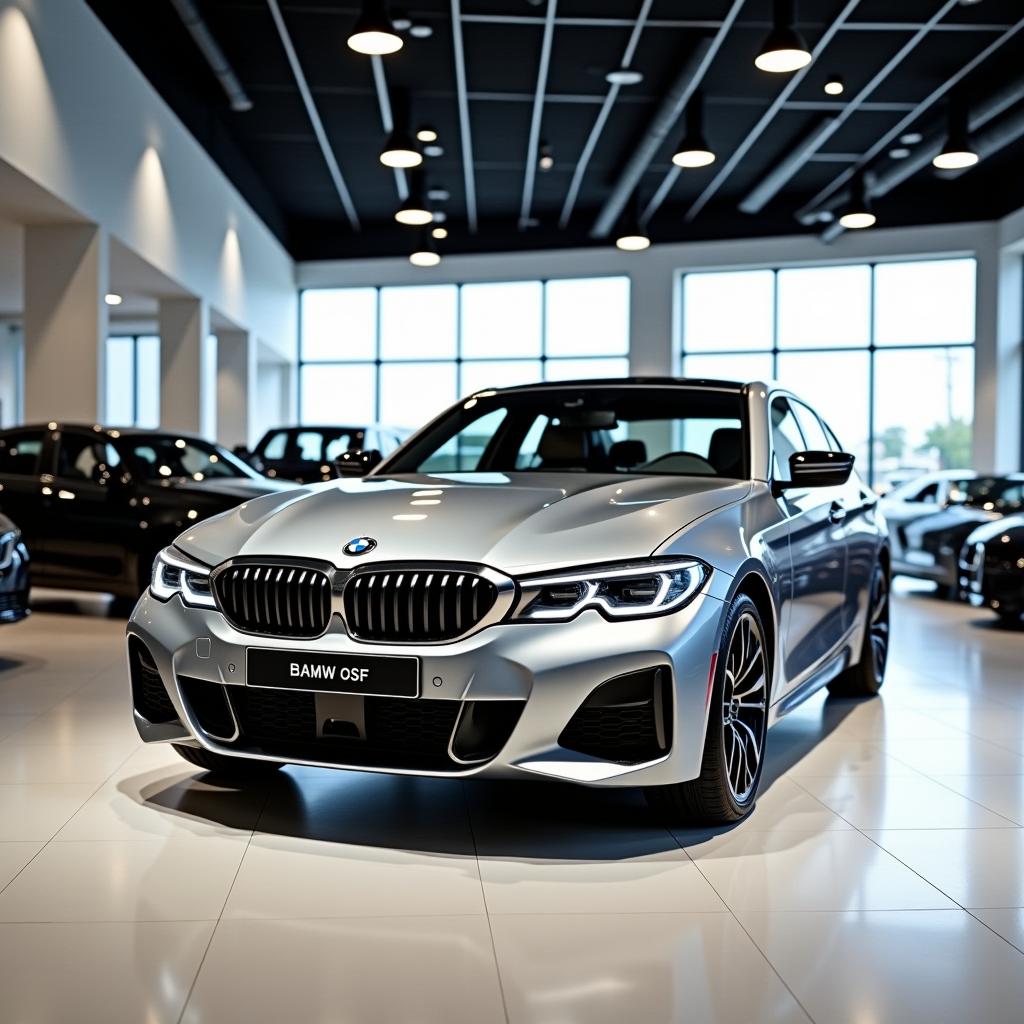} &
        \includegraphics[width=0.12\textwidth]{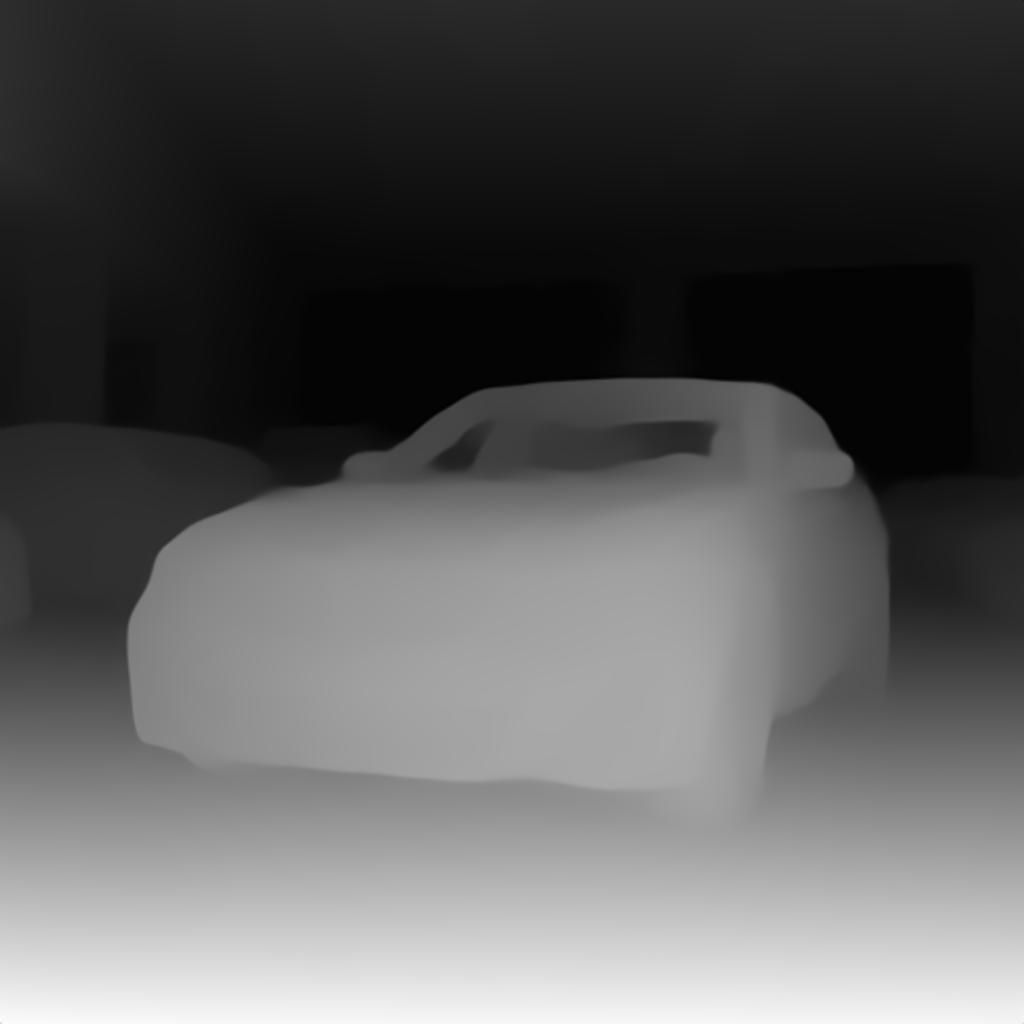} &
        \includegraphics[width=0.12\textwidth]{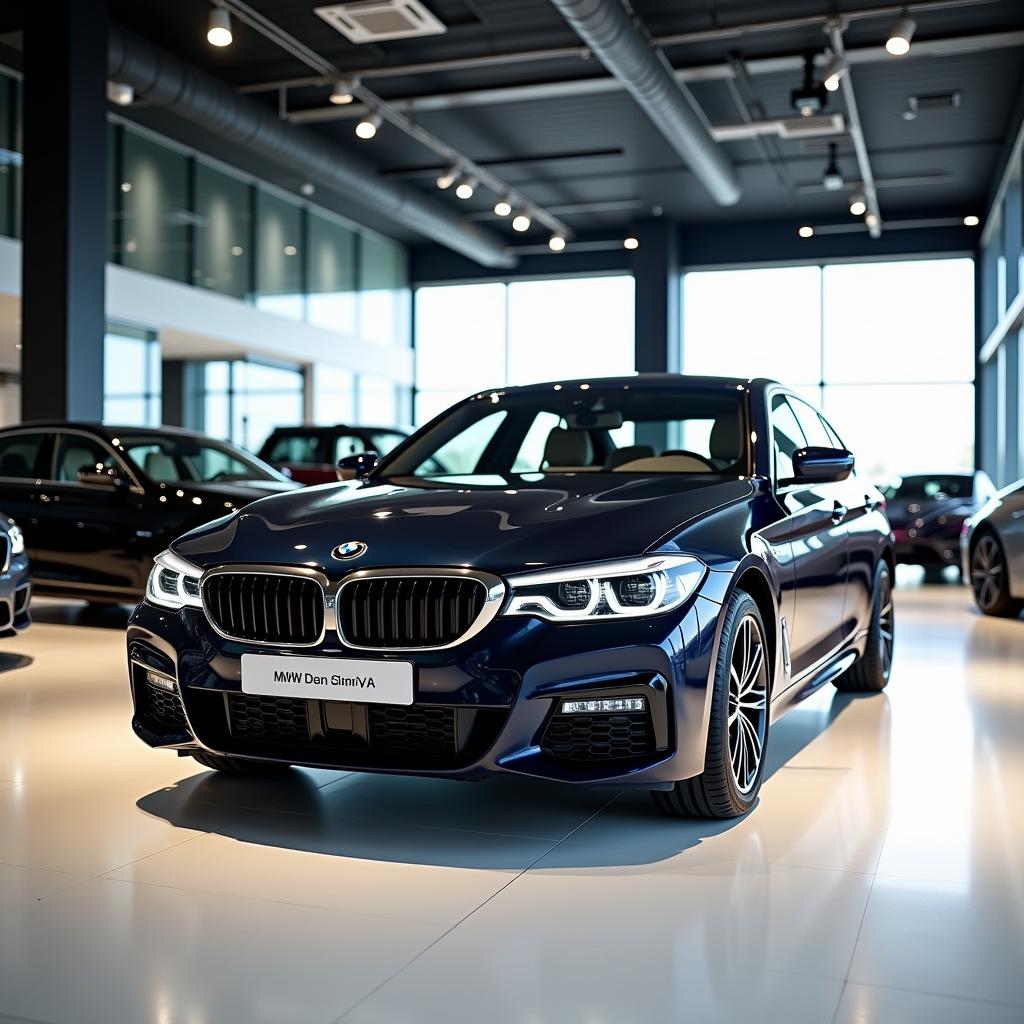} &
        \includegraphics[width=0.12\textwidth]{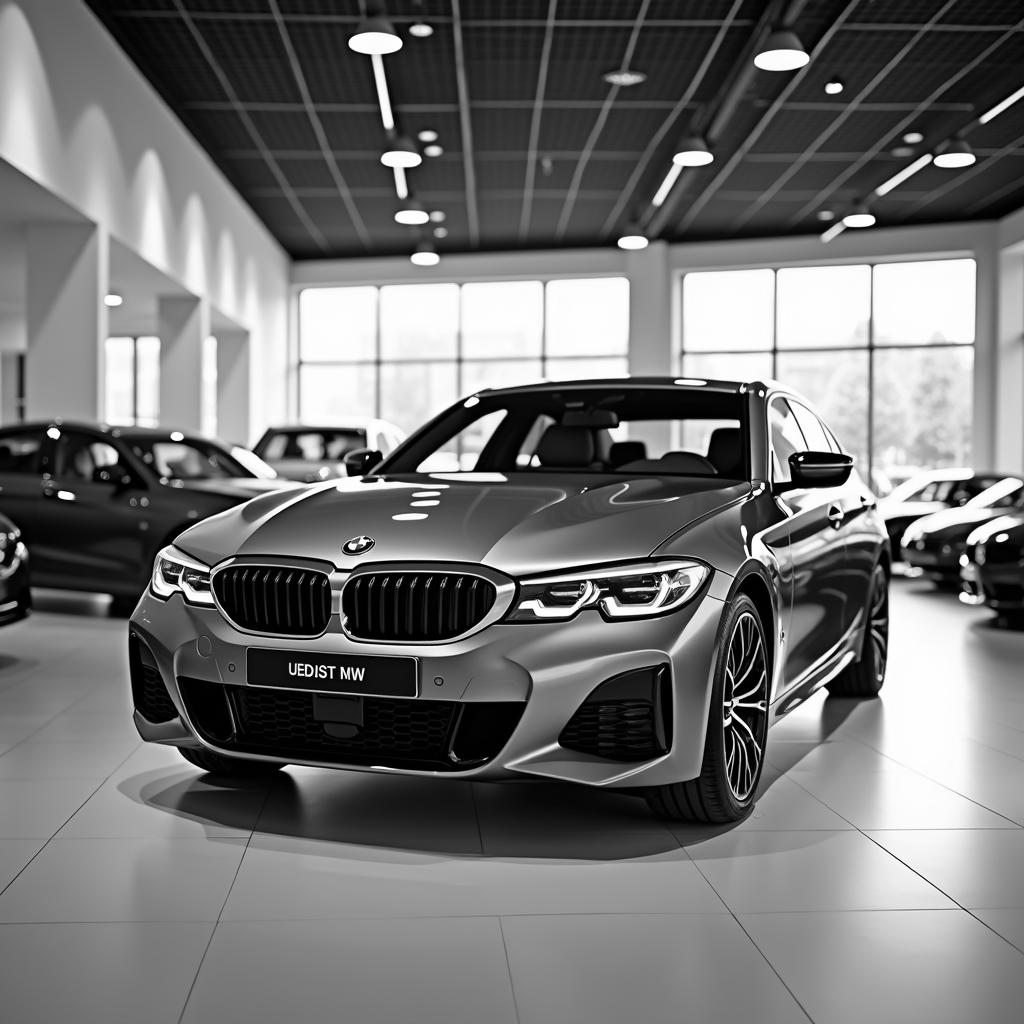} &
        \includegraphics[width=0.12\textwidth]{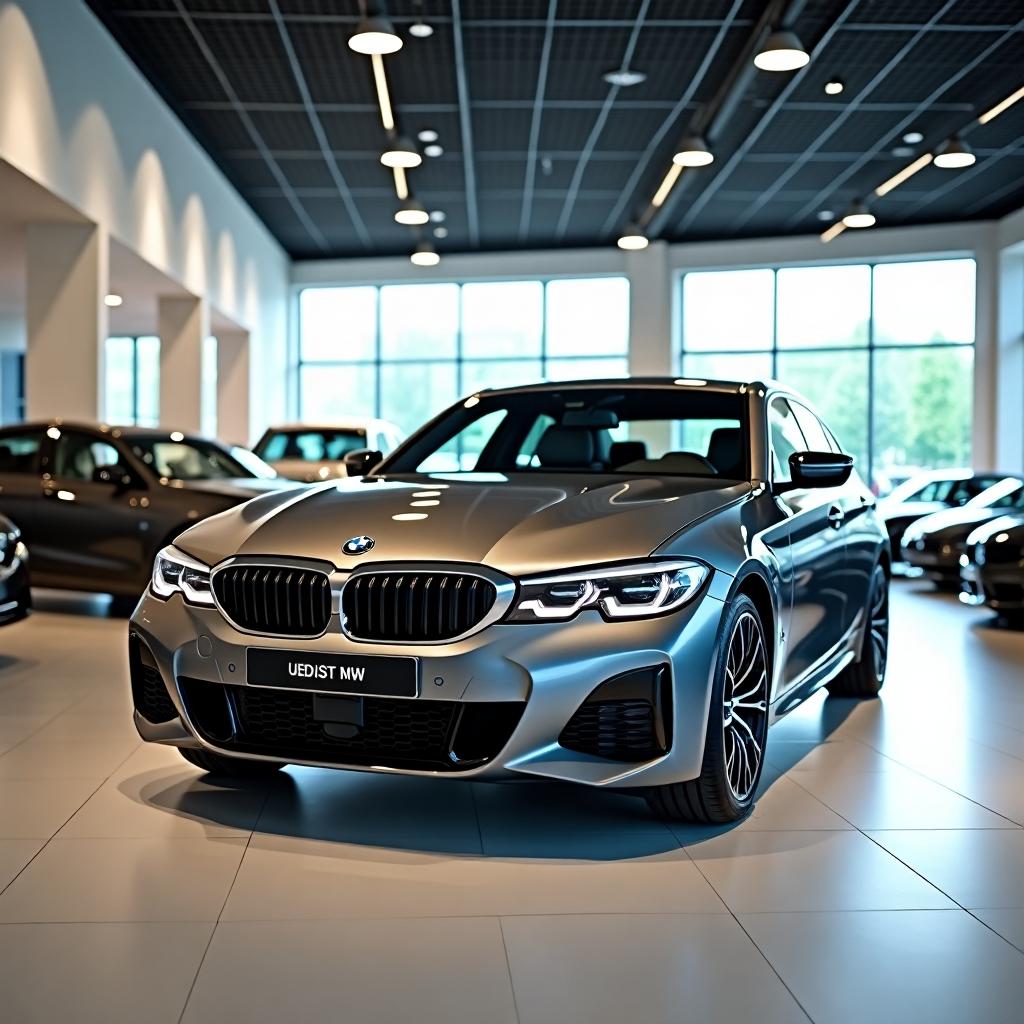} &
        \includegraphics[width=0.12\textwidth]{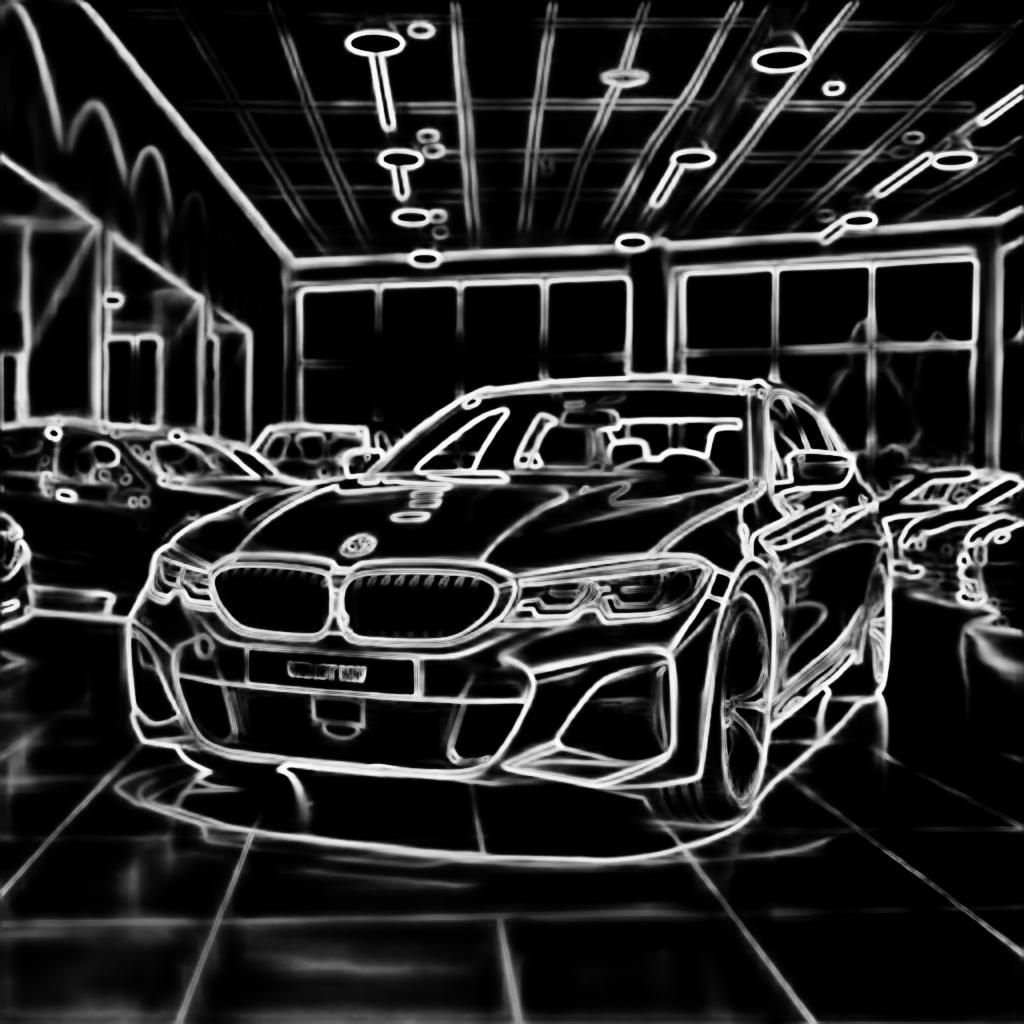} &
        \includegraphics[width=0.12\textwidth]{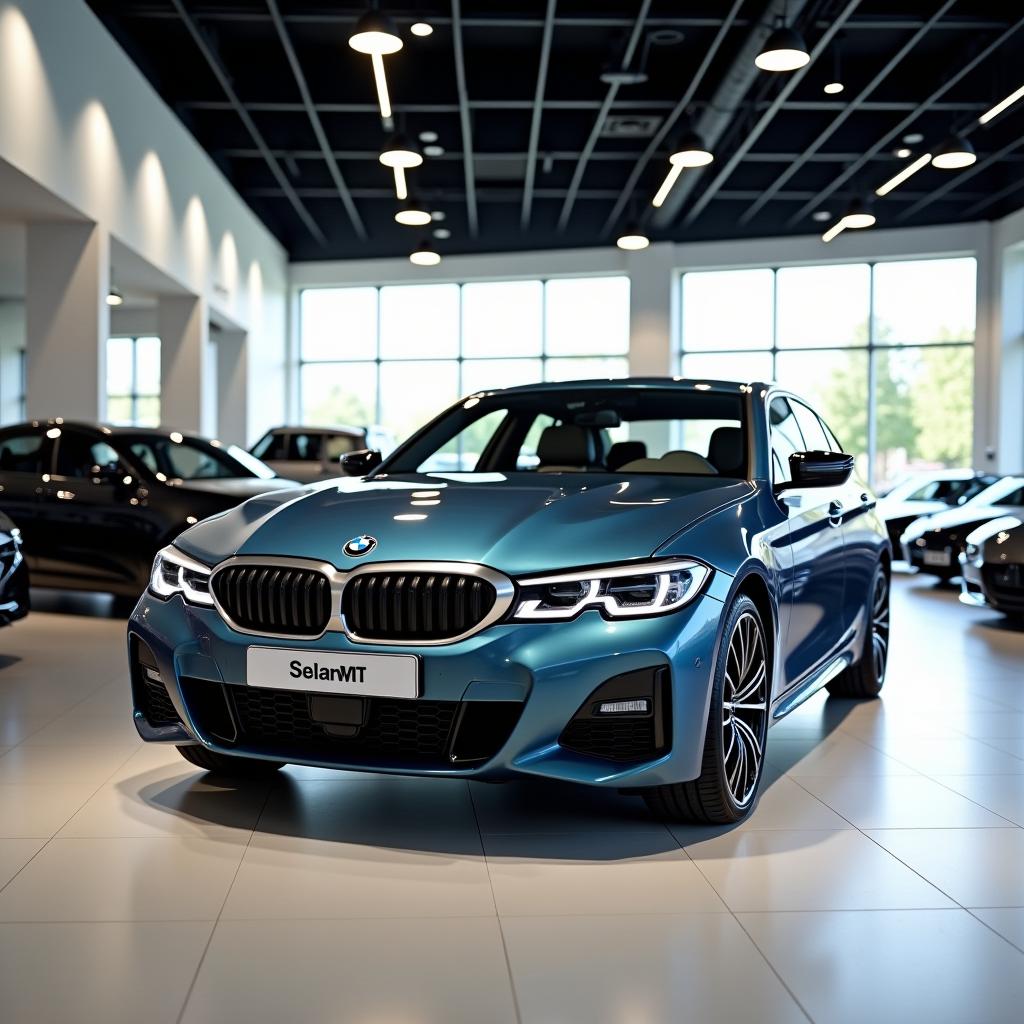} \\
        \multicolumn{8}{l}{\scriptsize \parbox[t]{0.95\textwidth}{A BMW car parked in a showroom, showcasing its sleek design and luxurious appearance. The car is positioned in the center of the scene, occupying a significant portion of the image. The showroom is well-lit, creating an inviting atmosphere for potential buyers. The car is displayed prominently, highlighting its features and attracting attention to the brand.}} \\
    
        
    \end{tabular}
    \caption{More visualized results on the four tasks, where the top two rows show the results at 512 resolution and the bottom two rows show the results at 1024 resolution.}
    \label{fig:more_visual}
\end{figure*}

Figure \ref{fig:more_visual} presents additional visual results of our method on the four tasks, including outputs at both 512 and 1024 resolutions.

\end{document}